\documentclass[conference]{IEEEtran}
\IEEEoverridecommandlockouts
\usepackage{amsmath,amsfonts}
\usepackage{array}
\usepackage[caption=false,font=normalsize,labelfont=sf,textfont=sf]{subfig}
\usepackage{textcomp}
\usepackage{stfloats}
\usepackage{url}
\usepackage{verbatim}
\usepackage{graphicx}
\usepackage{cite}

\newcommand{\bX}{\boldsymbol{X}}
\newcommand{\bx}{\boldsymbol{x}}
\newcommand{\bF}{\boldsymbol{F}}

\newcommand{\bC}{\boldsymbol{C}}
\newcommand{\bc}{\boldsymbol{c}}

\newcommand{\bff}{\boldsymbol{f}}
\newcommand{\bS}{S}

\newcommand{\myparagraph}[1]{\vspace{0.3\baselineskip}\noindent{\textbf{#1.}}~}

\newcommand{\BCLS}{\texttt{BCLS}\xspace}

\newcommand{\CDKM}{\texttt{CDKM}\xspace}
\newcommand{\BCD}{\texttt{BCD}\xspace}
\newcommand{\Lloyd}{\texttt{Lloyd}\xspace}
\newcommand{\DPFMPS}{\texttt{DPFMPS}\xspace}
\newcommand{\VFLC}{\texttt{VFLC}\xspace}
\newcommand{\FCE}{\texttt{FCE}\xspace}
\newcommand{\Fabric}{\texttt{Fabric}\xspace}
\newcommand{\Tebmean}{\texttt{Teb}\xspace}
\newcommand{\FabricNoDP}{\texttt{Fabric-NoDP}\xspace}

\newcommand{\Diabetes}{\textsf{Diabetes}\xspace}

\newcommand{\Census}{\textsf{Census}\xspace}
\newcommand{\Student}{\textsf{Student}\xspace}

\newcommand{\seconddata}{\textsf{Census1990}\xspace}
\newcommand{\HMDA}{\textsf{HMDA}\xspace}
\newcommand{\Letter}{\textsf{Letter}\xspace}
\newcommand{\Loan}{\textsf{Loan}\xspace}
\newcommand{\Patient}{\textsf{Patient}\xspace}
\newcommand{\Taxi}{\textsf{Taxi}\xspace}
\newcommand{\YearPredictionMSD}{\textsf{YearPrediction}\xspace}

\usepackage{booktabs}
\usepackage{pifont}
\newcommand{\cmark}{\ding{51}}  
\newcommand{\xmark}{\ding{55}}

\usepackage{subcaption} 

\usepackage{amsthm}

\newtheoremstyle{mydefinition} 
  {1pt} 
  {1pt} 
  {\itshape} 
  {} 
  {\bfseries} 
  {.} 
  { } 
  {} 

\theoremstyle{mydefinition}

\newtheorem{theorem}{Theorem}

\newtheorem{definition}{Definition}

\usepackage{mathtools}

\usepackage{amsmath}
\usepackage{amsthm}
\usepackage{enumitem}
\usepackage{multirow}

\usepackage{xcolor}

\usepackage{ragged2e}
\usepackage{etoolbox}
\usepackage{xcolor}
\usepackage{algorithmic}
\usepackage[linesnumbered,ruled,vlined,boxed,noend]{algorithm2e}

\usepackage{graphicx}

\usepackage[caption=false,font=normalsize,labelfont=sf,textfont=sf]{subfig}

\usepackage[
  colorlinks=true,
  linkcolor=black,
  citecolor=blue,
  urlcolor=blue
]{hyperref}

\begin{document}

\title{Federated and Balanced $k$-means: Communication Efficiency, Privacy, and Heterogeneity}

\IEEEoverridecommandlockouts

\author{
\IEEEauthorblockN{
Feiteng Nie\textsuperscript{1$\dagger$},
Shengkun Zhu\textsuperscript{1$\dagger$},
Sheng Wang\textsuperscript{1*},
Yuan Sun\textsuperscript{2},
Feiping Nie\textsuperscript{3},
Wei Song\textsuperscript{1},
Zhiyong Peng\textsuperscript{1}
}

\IEEEauthorblockA{\textsuperscript{1}School of Computer Science, Wuhan University}
\IEEEauthorblockA{\textsuperscript{2}La Trobe Business School, La Trobe University}
\IEEEauthorblockA{\textsuperscript{3}School of Artificial Intelligence, Northwestern Polytechnical University}

\IEEEauthorblockA{
\{niefeiteng, whuzsk66, swangcs, songwei, peng\}@whu.edu.cn\\
yuan.sun@latrobe.edu.au; feipingnie@gmail.com
}

\thanks{\textsuperscript{*} Sheng Wang is the corresponding author.}
\thanks{\textsuperscript{$\dagger$} Feiteng Nie and Shengkun Zhu contributed equally to this work.}
}






\maketitle

\begin{abstract}
Collaborating on data across organizations in healthcare and finance demands methods that preserve privacy while enabling effective analysis.
Vertical Federated Learning (VFL) facilitates clustering across data where different parties hold different features of the same entities without sharing raw data.  
However, existing VFL clustering methods suffer from two key limitations: vulnerability to information leakage and the creation of imbalanced clusters. 
To address these limitations, we propose \Fabric, a balanced and privacy-preserving federated $k$-means algorithm. 
Our method decomposes the balanced clustering problem into independent subproblems for each client, using the Alternating Direction Method of Multipliers (ADMM) to efficiently handle the balance constraints. 
Under the semi-honest model, \Fabric provides provable privacy guarantees by injecting calibrated noise into local objective evaluations.
Extensive evaluations on ten real-world datasets show that \Fabric achieves a superior utility-privacy trade-off compared with state-of-the-art methods and converges faster.
Specifically, \Fabric reduces SSE by 29.1\% on average and improves balance performance by 1.41$\times$, while being 26$\times$ faster on average.
\end{abstract}


\section{Introduction}
With the rapid development of information technology, clustering has become a crucial tool in data analysis, particularly for processing and interpreting large datasets~\cite{Wang20Efficiency}.
$k$-means serves as a fundamental clustering method across diverse domains, including healthcare \cite{Huang2022OVQA, mateo2008balanced, Tan2022MetaCare}, finance \cite{chandramouli2006database, Arash2020Vertica, Zhou2009Graph}, and education \cite{ZhangW0P23, He2022Unsupervised}. 
It helps reveal latent patterns in data and facilitates effective data-driven decision-making.

\begin{figure}[t]
  \centering
  \includegraphics[width=\linewidth]{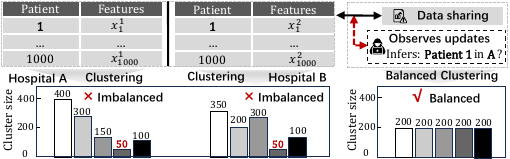}
  \vspace{-1.5em}
  \caption{
  Clustering without data sharing leads to inaccurate results and clinically infeasible small clusters.
  Enabling data sharing and balanced clustering can achieve equal cluster sizes, but still requires privacy protection measures.
}
  \vspace{-0.95em} 
  \label{fig: Example1}
\end{figure}

However, applying $k$-means in these domains is challenging, as data are distributed across institutions and legal and regulatory constraints limit data sharing.
Consider the scenario in Fig. \ref{fig: Example1}: two hospitals clustering 1,000 chronic obstructive pulmonary disease (COPD) patients to identify disease subtypes for drug trials. 
The two hospitals possess complementary patient features: Hospital A holds demographics, and Hospital B possesses pulmonary function tests. 
Accurate clustering requires jointly analyzing these data, but strict data protection regulations such as HIPAA \cite{Annas2003HIPAA} and GDPR \cite{Hoofnagle2019GDPR} prohibit direct data sharing across institutions, thus requiring each hospital to cluster independently using only local features. 
If hospitals can access additional patient features while complying with regulations, their ability to analyze and classify disease subtypes will improve significantly.
Vertical Federated Learning (VFL) provides a framework for enabling parties to perform collaborative clustering without data sharing~\cite{Wu2026Trustworthy,wang2026defense, kou2026fedharmony,zhou2025hounding}.

While VFL enables collaboration across institutions, directly applying federated $k$-means introduces two major problems: imbalanced results and privacy risks.
First, as shown in Fig. \ref{fig: Example1}, $k$-means may produce highly imbalanced clusters, such as clusters containing 400, 300, 150, 50, and 100 samples. 
A cluster with only 50 patients (5\%) is clinically unviable because it fails to meet the minimum sample size requirements for adequately powered clinical trials~\cite{cohen2013statistical, guidance2018master}. 
In contrast, balanced clusters (e.g., 200 patients per subtype) yield clinically viable subtypes.
Second, VFL does not provide a formal privacy guarantee. 
An attacker may infer whether Patient 1 is at Hospital A, which constitutes a membership inference attack \cite{Shokri17Membership}. 
Therefore, VFL should be combined with additional privacy mechanisms, such as differential privacy (DP) \cite{dwork2004privacy, li2018differentially, shi2022distributionally}, to provide provable privacy protection.

Balanced clustering is a key focus of our work.
Specifically, balanced clustering aims to partition data into clusters with approximately equal size~\cite{ding2002cluster, zhou2020unsupervised}, 
which is crucial in applications such as healthcare~\cite{mateo2008balanced}, retrieval systems~\cite{althoff2011balanced}, and index design~\cite{chen2021spann}.
Existing balanced clustering can be categorized into hard-balanced clustering and soft-balanced clustering. 
Hard-balanced clustering formulates clustering as a constrained assignment problem (e.g., via the Hungarian algorithm in BKM~\cite{Malinen14Balanced}), thereby guaranteeing strict balance at the cost of high combinatorial complexity and poor scalability.
Soft-balanced clustering adds balance penalty terms to the objective function~\cite{Liu17Balanced, Li18Balanced, Lin19BalancedClustering}, but introduces additional hyperparameters and sensitivity to their tuning.
However, these methods assume centralized access to all features, which is unrealistic in practice because the features are distributed across institutions where direct data sharing is not allowed.
This motivates the exploration of balanced clustering that does not require direct data sharing between institutions.

\begin{table}[t]
\centering
\caption{Comparison of different methods}
\vspace{-0.5em} 
\label{tab:method_comparison}
\resizebox{\linewidth}{!}{
\begin{tabular}{lcccc}
\toprule
\textbf{Methods} & \textbf{Balance} & \textbf{VFL} & \textbf{Privacy-Preserving} & \textbf{Time Complexity} \\
\midrule
\CDKM~\cite{nie2022CDKM} & \xmark & \xmark & \xmark & $O(ndkT)$  \\
\BCLS~\cite{Liu17Balanced} & \cmark & \xmark & \xmark & $O(n^2kT + d^T)$  \\
\FCE~\cite{zhou2024fair} & \cmark & \xmark & \xmark &  $O(n^2k^2)$\\
\VFLC~\cite{Ding16KMeans} & \xmark & \cmark & \xmark & $O(ndk + dk^{H})$ \\
\DPFMPS~\cite{li2022differentially} & \xmark & \cmark & \cmark & $O(ndk + dk^{H})$ \\
\textbf{\Fabric} & \cmark & \cmark & \cmark & $O(ndkT)$ \\
\bottomrule
\end{tabular}
}
\vspace{-1.0em}
\end{table}

Privacy preservation is an important requirement in collaborative data analytics. 
However, VFL cannot provide any provable privacy guarantee in the sense that adversaries can still infer whether a patient is in the hospital by examining the shared information from local parties~\cite{hong2014collaborative, wang2019beyond, squicciarini2014privacy}.
Moreover, existing VFL methods struggle to scale to large numbers of clients, and their reliance on exchanging intermediate statistics (e.g., cluster centroids) raises serious privacy concerns.
VFLC~\cite{Ding16KMeans} developed a communication-efficient $k$-means variant, but its dependence on Cartesian-product matching across parties makes it hard to scale.
MPC-based approaches~\cite{vaidya2003privacy} incur high computational costs while providing privacy.
\DPFMPS~\cite{li2022differentially} provides DP~\cite{Dwork06Calibrating, Erlingsson@2014RAPPOR} but requires sharing cluster centroids and does not scale well to a large number of clients.

As summarized in Table \ref{tab:method_comparison}, no prior work simultaneously addresses balance, vertical federation, and privacy protection within a unified and effective framework.
This limitation mainly stems from the inherent tension between these objectives: balance constraints and privacy preservation often degrade clustering quality, while federated optimization further constrains the design space for maintaining utility.

In this paper, we propose \Fabric to address the limitations of existing methods in the VFL framework.
We first reformulate the balanced clustering problem using inequality-based constraints within an ADMM framework, which naturally decomposes into subproblems suitable for clients to solve. 
By leveraging block coordinate descent, clients exchange only optimization results rather than raw data or cluster centers, thereby achieving both communication efficiency and enhanced privacy protection. 
We further incorporate DP to provide formal privacy guarantees against inference attacks.
Our main contributions are as follows:
\begin{itemize}   
\item 
We propose an efficient ADMM-based method for balanced $k$-means in VFL that achieves a significant reduction in communication rounds (see Section \ref{sec: subproblem Fabric means}).
\item 
We integrate DP into \Fabric to provide formal privacy guarantees with negligible overhead on clustering utility and efficiency (see Section~\ref{sec: Fabric means DP}).
\item 
We establish that \Fabric is robust to feature heterogeneity across clients and provides stronger privacy guarantees by eliminating cluster center exchange (see Section~\ref{sec: Discussions}).
\item 
Extensive experiments on ten real-world datasets demonstrate that \Fabric outperforms state-of-the-art methods in utility, balance, and privacy protection, while achieving an average speedup of 26$\times$ (see Section \ref{sec: experiments}).
\end{itemize}

\section{Related Work}
In this section, we review related work on balanced clustering, federated clustering, and privacy-preserving clustering.

\noindent\textbf{Balanced clustering.}
In recent years, balanced clustering has been extensively studied~\cite{Chen05Size, Malinen14Balanced, Liu18FastClustering, Liu17Balanced, Li18Balanced, Lin19BalancedClustering, Xu24Semi, nie2024parameter}.
Chen et al. \cite{Chen05Size} incorporated prior knowledge of cluster sizes into the clustering process through size-regularized partitioning.
BKM \cite{Malinen14Balanced} employed the Hungarian algorithm for balanced assignment, but this increased the time complexity to $\mathcal{O}(n^3)$.
Several methods~\cite{Liu17Balanced, Li18Balanced, Lin19BalancedClustering} introduced penalty terms to the objective function to achieve balance. 
BCLS \cite{Liu17Balanced} applied the augmented Lagrangian multiplier method to regularize the clustering model through least-squares regression.
Li et al. \cite{Li18Balanced} further optimized the objective using exclusive lasso to obtain more balanced partitions.
Lin et al. \cite{Lin19BalancedClustering} introduced a uniform balance model and replaced the assignment step with a simplex-based optimization procedure.
Liu and Huang et al. \cite{Liu18FastClustering} proposed FCFC, which combines the standard $k$-means objective with an imbalance penalty based on cluster-size variance.
BKHK \cite{Nie2022Unsupervised} is an efficient balance-based hierarchical $k$-means method, but the final number of clusters is restricted to powers of two.
Zhou et al. \cite{zhou2024fair} added a regularization term to enforce equal cluster sizes.
However, these existing approaches often suffer from suboptimal clustering utility, high sensitivity to penalty parameters, and significant computational overhead (as summarized in Table~\ref{tab:method_comparison}). 

\noindent\textbf{Federated clustering.}
Cluster analysis in VFL has gained significant attention \cite{Ding16KMeans, li2022differentially, Zhu23F3KM, ji2025federated}. 
VFLC~\cite{Ding16KMeans} developed a communication-efficient variant of $k$-means in which each client computes local centers using its own feature subset, and the server aggregates these local centers.
DPFMPS~\cite{li2022differentially} introduced a differentially private $k$-means approach in which the server constructs a weighted grid to summarize the global dataset.
However, constructing weighted grids incurs a computational complexity of $\mathcal{O}(k^H)$ (as shown in Table~\ref{tab:method_comparison}).
Huang et al. \cite{Huang22Coresets} developed a framework for constructing coresets in VFL settings and applied it to $k$-means. 
Zhu et al. \cite{Zhu23F3KM} proposed an in-processing method to solve the subproblems of federated fair $k$-means.
\Tebmean~\cite{ji2025federated} adopted a penalty method for balance constraints.
Although penalty-based formulations are easy to implement, they only enforce balance in a soft manner and require careful tuning of penalty parameters; otherwise, clustering utility and balance quality may vary substantially.
In contrast, our method introduces auxiliary variables and uses ADMM to directly handle explicit cluster-size bounds, which provides more controllable balance constraints while preserving the decomposability needed in VFL.
Moreover, existing VFL clustering methods based on Cartesian product-based weighted grids also incur high computational complexity.

\noindent\textbf{Privacy-preserving clustering.}
Various methods have been developed to provide DP guarantees for $k$-means.
Blum et al. \cite{Blum05Practical} first introduced differentially private $k$-means by adding Laplace noise at each iteration.
Mohan et al. \cite{Mohan12GUPT} developed a DP analysis system based on sampling and aggregation.
Su et al. \cite{Su16Differentially}, Balcan et al. \cite{Balcan17Differentially}, and Huang et al. \cite{huang2018optimal} proposed efficient implementations of differentially private $k$-means. 
Lu et al. \cite{Lu21Differentially} controlled the movement direction of centroids during iterations to support convergence.
Yang et al. \cite{Yang22kMeans} proposed a local privacy-preserving $k$-means algorithm, which employs a bounded perturbation mechanism and formally proves that the mechanism satisfies DP for Euclidean distance. 
Li et al. \cite{li2022differentially} constructed weighted grids to derive global cluster centers with DP guarantees in VFL. 
However, because the grid nodes are constructed as a Cartesian product of local centers, scalability becomes a serious issue as the number of clients increases. 
In addition, noise is added to the centers to preserve privacy, which impacts clustering utility.
Zhu et al. \cite{Zhu24Privacy} proposed a privacy-preserving federated learning method over vertically partitioned datasets using fuzzy clustering and fuzzy rule models.

\noindent\textit{\underline{\noindent\textbf{Remarks.}}}
Existing methods achieve notable results on individual aspects, yet none simultaneously address balance, vertical federation, and privacy protection in a unified framework (see Table~\ref{tab:method_comparison}).
This gap stems from the inherent tension between these objectives: enforcing balance and privacy often degrades clustering quality, while the requirement of efficient federated optimization further limits the scope for preserving utility.

\section{Preliminaries}

In this section, we present the $k$-means formulation of Nie et al. \cite{nie2022CDKM}, which reformulates the traditional $k$-means objective function into a centerless version. 
We then provide detailed definitions of ADMM and DP. 
Table~\ref{tab: notations} summarizes the notations used in this paper.

\subsection{Problem Formulation of $k$-means}
\label{sec:problem_formulation}
The $k$-means clustering problem aims to partition $n$ data points,  represented by $\bX \in \mathbb{R}^{d \times n}$, into $k$ clusters by minimizing the sum of squared errors (SSE):
\begin{align}
\small\label{eq:sse_basic}
\min_{\mathcal{S}, \bC} \sum_{j=1}^{k} \sum_{\bx_i \in \bS_j} \|\bx_i - \bc_j\|_2^2,
\end{align}
where $\mathcal{S} = \{\bS_1, \ldots, \bS_k\}$ denotes the cluster partition and $\bc_j$ is the centroid of cluster $\bS_j$. 
Recently, Nie et al.~\cite{nie2022CDKM} reformulated the $k$-means objective as a trace maximization problem with the indicator matrix $\bF$ as the sole optimization variable, thereby eliminating the need to compute cluster centroids explicitly. 
Specifically, the indicator matrix $\bF \in \mathbb{R}^{n \times k}$ is defined such that $f_{ij} = 1$ if point $\bx_i$ belongs to cluster $\bS_j$, and $f_{ij} = 0$ otherwise. 
Each row of $\bF$ contains exactly one non-zero element, ensuring each data point is assigned to exactly one cluster.
Based on this representation, Problem~\eqref{eq:sse_basic} can be reformulated as follows:
\begin{align}\label{eq:sse_matrix}
\small
\min_{\bF \in \text{Ind}, \bC} \sum_{j=1}^{k}\sum_{i=1}^{n} \|\bx_i - \bc_j\|_2^2 f_{ij} = \min_{\bF \in \text{Ind}, \bC} \|\bX - \bC\bF^\top\|_F^2,
\end{align}
where $\|\cdot\|_F$ denotes the Frobenius norm. Problem \eqref{eq:sse_matrix} can be reformulated as a trace minimization problem:
\begin{align}\label{eq:sse_trace}
\small
\min_{\bF \in \text{Ind}, \bC} \text{Tr}\left((\bX - \bC\bF^\top)(\bX - \bC\bF^\top)^\top\right).
\end{align}
Taking the derivative of~\eqref{eq:sse_trace} with respect to $\bC$ and setting it to zero yields the closed-form solution for the optimal centroids:
\begin{align}\label{eq:optimal_C}
\small
\bC = \bX\bF(\bF^\top\bF)^{-1}.
\end{align}
Substituting~\eqref{eq:optimal_C} into~\eqref{eq:sse_trace} gives:
\begin{align}\label{eq:trace_min}
\small
\min_{\bF \in \text{Ind}} -\text{Tr}\left((\bF^\top\bF)^{-1}\bF^\top\bX^\top\bX\bF\right).
\end{align}

Since each row of $\bF$ contains only one nonzero entry, the matrix $\bF^\top\bF$ is diagonal, and its $j$-th diagonal entry is given by $\bff_j^\top \bff_j$, where $\bff_j$ is the $j$-th column of $\bF$.
Therefore, Problem~\eqref{eq:trace_min} can be equivalently rewritten as
\begin{align}\label{eq:obj_f_final}
\small
\max_{\bF \in \text{Ind}} \text{obj}(\bF) = \max_{\bF \in \text{Ind}} \sum_{j=1}^{k} \frac{\bff_j^\top \bX^\top \bX \bff_j}{\bff_j^\top \bff_j}.
\end{align}
Therefore, solving Problem~\eqref{eq:sse_basic} is equivalent to solving Problem~\eqref{eq:obj_f_final} with one variable $\bF$.

\begin{table}
\centering
\small
\caption{Summary of notations}
\vspace{-0.5em}
\label{tab: notations}
\resizebox{\linewidth}{!}{
\begin{tabular}{ll}
\toprule
\textbf{Notation} & \textbf{Description} \\
\midrule
$n, d \in \mathbb{N}^+$ & The dataset size and dimension \\
$\bX \in \mathbb{R}^{d \times n}$ & The dataset  \\
$\mathcal{S} = \{\bS_1, \bS_2, \cdots, \bS_k\}$ & The set of $k$ clusters \\
$\textbf{C} \in \mathbb{R}^{d \times k}$ & The center matrix \\
$\bF \in \mathbb{R}^{n \times k}$ & The indicator matrix \\
$\boldsymbol{\Theta}_h \in \mathbb{R}^{L_h \times k}$ & Auxiliary variable in client h\\
$\boldsymbol{\lambda}_h \in \mathbb{R}^{L_h \times k}$ & Dual variable in client h\\
$\boldsymbol{\alpha}, \boldsymbol{\beta} \in \mathbb{R}^{k}$ & The bounds of  cluster size\\
$\epsilon \in \mathbb{R}^+$ & Privacy budget \\
\bottomrule
\end{tabular}
}
\vspace{-1em}
\end{table}

\subsection{Alternating Direction Method of Multipliers}
For the balance constraints arising in clustering, we employ the alternating direction method of multipliers (ADMM) \cite{Stephen2011Distributed, zhu2024admm}, an optimization method
suited for the following problem:
\begin{equation}
\small
\begin{aligned}
\label{eq:admm_standard_form}
\min_{\mathbf{x},\mathbf{z}}\quad 
& f(\mathbf{x}) + g(\mathbf{z}) 
\quad \text{s.t.}\quad 
& A\mathbf{x} + B\mathbf{z} = \mathbf{c},
\end{aligned}
\end{equation}
where $\mathbf{x}\in\mathbb{R}^n$, $\mathbf{z}\in\mathbb{R}^m$,
$A\in\mathbb{R}^{p\times n}$, $B\in\mathbb{R}^{p\times m}$,
and $\mathbf{c}\in\mathbb{R}^p$.
The augmented Lagrangian is given by:
\begin{equation}
\small
\label{eq:augmented_lagrangian}
\mathcal{L}_\rho(\mathbf{x},\mathbf{z},\mathbf{y})
=f(\mathbf{x}) + g(\mathbf{z})
+ \mathbf{y}^\top(A\mathbf{x} + B\mathbf{z} - \mathbf{c})
+ \frac{\rho}{2}\big\|A\mathbf{x} + B\mathbf{z} - \mathbf{c}\big\|_2^2,
\end{equation}
where $\mathbf{y}\in\mathbb{R}^p$ denotes the Lagrange multiplier and $\rho>0$ is the penalty parameter.
ADMM iterations then consist of the following alternating updates:
\begin{equation}
\small
\label{eq:admm_iterations}
\begin{cases}
\mathbf{x}^{t+1}
= \displaystyle\arg\min_{\mathbf{x}}\ 
\mathcal{L}_\rho(\mathbf{x},\mathbf{z}^t,\mathbf{y}^t), \\
\mathbf{z}^{t+1}
= \displaystyle\arg\min_{\mathbf{z}}\ 
\mathcal{L}_\rho(\mathbf{x}^{t+1},\mathbf{z},\mathbf{y}^t), \\
\mathbf{y}^{t+1}
= \mathbf{y}^t
+ \rho\big(A\mathbf{x}^{t+1} + B\mathbf{z}^{t+1} - \mathbf{c}\big).
\end{cases}
\end{equation}
where $t$ denotes the iteration index.
ADMM offers efficient distributed and parallel optimization, handles equality constraints via an augmented Lagrangian, and enjoys global convergence guarantees in the convex setting~\cite{Stephen2011Distributed}, making it particularly suitable for machine learning.
These convergence guarantees apply to convex problems. 
In the balanced $k$-means setting considered in this paper, the assignment variables lead to a discrete and nonconvex optimization problem.
Hence, ADMM is used as an efficient local solver rather.


\subsection{Differential Privacy}
Differential privacy \cite{Dwork06Calibrating} aims to prevent attackers from identifying data from any specific individual.
To achieve this, it adds carefully calibrated noise, ensuring that the presence or absence of any single individual does not significantly affect the outcome.
Formally, we use the following standard definition:
\begin{definition}[Differential Privacy \cite{Dwork06Calibrating}]\label{def:dp}
Let $\epsilon, \delta \in \mathbb{R}^+$. 
A randomized algorithm $\mathcal{M}$ with output range $\mathcal{R}$ satisfies $(\epsilon,\delta)$-DP if, for any two neighboring datasets $\bX$ and $\bX'$ that differ in one individual record and for any output event $\mathcal{O}\subseteq\mathcal{R}$, the following inequality holds:
\begin{align}\label{eq:DP definition}
\begin{aligned}
\Pr[\mathcal{M}(\bX) \in \mathcal{O}] \leq e^{\epsilon} \Pr[\mathcal{M}(\bX') \in \mathcal{O}] + \delta,
\end{aligned}
\end{align}
where $\epsilon$ is the privacy budget and $\delta$ is a relaxation term.
\end{definition}
DP ensures that the inclusion or exclusion of any single individual's data has a negligible impact on the algorithm's output, thereby safeguarding individual privacy.

\section{ALGORITHM DESIGN}
In this section, we first formalize the privacy-preserving balanced clustering in VFL. 
We then discuss the challenges of the problem and introduce our method: \Fabric.
Finally, we discuss the advantages of our methods compared to other VFL clustering frameworks and present a complexity analysis.

\subsection {Problem Formulation}
We formalize the vertical federated balanced clustering problem and specify our threat model.

\myparagraph{Vertical federated balanced clustering}
Balanced $k$-means aims to assign all samples to $k$ clusters with equal cluster sizes~\cite{Liu17Balanced}.
Formally, the size of each cluster $j$ is bounded by $\alpha_j \leq \left| \{ \bx_i \mid \bx_i \in \bS_j \} \right| \leq \beta_j$, where $ \bS_j $ denotes the set of samples assigned to cluster $j$.
Using the indicator matrix $\bF$, where $ \bff_j^{\top} \bff_j$ equals the number of samples in cluster $j$, we can express the balance constraint as:
\begin{align}
\small
\alpha_j \leq \bff_j^{\top} \bff_j \leq \beta_j, \quad j \in [k].
\end{align}
We reformulate \eqref{eq:obj_f_final} with balance constraints as follows:
\begin{equation}
\small
\begin{aligned}\label{eq: clustering with st}
\min_{\bF \in \textit{Ind}} - \sum_{j=1}^{k} \frac{\bff_j^{\top} \bX^{\top} \bX \bff_j}{\bff_j^{\top} \bff_j}, 
\quad  \text{s.t.} \; \alpha_j \leq \bff_j^{\top} \bff_j \leq \beta_j, \; j \in [k].
\end{aligned}
\end{equation}
In the VFL setting, we can treat $\{\bX_h\}_{h=1}^{H}$ as a vertical partition of $\bX$ across $H$ clients, where each $\bX_h \in \mathbb{R}^{d_h \times n}$ contains a distinct subset of features for the same set of $n$ samples.
In this setting, each data point $\bx_i$ is partitioned as $\bx_i = (\bx_i^{(1)}; \bx_i^{(2)}; \cdots; \bx_i^{(H)})$, where $\bx_i^{(h)}$ denotes the features held by client $h$.
The clients collaborate with a central server to solve~\eqref{eq: clustering with st} in a federated manner. 
To preserve data privacy, each client computes local statistics from $\bX_h$ for server aggregation, ensuring raw features never leave the clients.

\myparagraph{Threat Model and Security Assumptions}
We adopt the honest-but-curious adversary model~\cite{goldreich2019play}, where:
\begin{itemize}
\item \textbf{Server:} Follows the protocol correctly but may attempt to infer private information from received messages (e.g., objective values, indicator matrix $\bF$).
    
\item \textbf{Clients:} Execute the algorithm faithfully but may try to deduce information about other clients' data by observing shared parameters \cite{truex2020ldp, miao2022compressed, liu2021flame}.
\end{itemize}
We assume that clients do not collude and that all communication channels are secured using standard cryptographic protocols. 
The primary threat considered in this work is membership inference attacks~\cite{Shokri17Membership}, where adversaries attempt to determine whether a specific point was used in clustering. 

\subsection{Key Challenges}
Achieving balanced clustering in VFL with privacy guarantees poses several fundamental challenges:
\begin{itemize}

\item How can we decompose the balanced clustering problem into subproblems that each client can solve independently without accessing the whole feature space?
 
\item Communication is a key bottleneck in VFL \cite{li2020federated}. How can we design communication-efficient updates when each iteration requires $n$ synchronization steps?

\item In VFL, clients typically hold different features with varying sizes, leading to heterogeneous feature distributions.
How can we handle such heterogeneity to ensure accurate and robust clustering remains a significant challenge?

\item How can we protect individual data points from inference attacks while maintaining clustering quality?
\end{itemize}

\myparagraph{Goal}
All clients aim to cooperatively generate balanced clustering results with $(\epsilon,\delta)$-DP without sharing raw data, while minimizing the k-means objective and maintaining communication efficiency in practical deployments.

\begin{figure}[t]
  \centering
  \includegraphics[width=\linewidth]{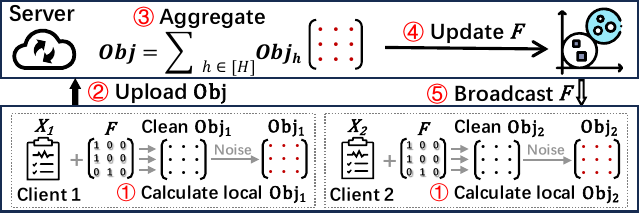}
  \vspace{-1.5em}
  \caption{Framework of \Fabric. Clients compute local objectives $Obj_h$ and upload them to the server. 
  The server aggregates, updates $\bF$, and broadcasts it to clients.}
  \vspace{-1.2em}
  \label{fig: framework fabricmeans}
\end{figure}

To address the above  challenges, we propose the following strategies: 
First, we decompose the balanced $k$-means objective into client-local subproblems with a lightweight consensus update (Section~\ref{sec: Fabric means}).
Second, each client solves its subproblem within ADMM using a block coordinate descent routine (\BCD) \cite{zeng2019BCD, Zhu23F3KM}, which reduces synchronization frequency and communication cost (Section \ref{sec: Fabric means}).
Third, to prevent potential leakage of client information, we incorporate $(\epsilon,\delta)$-DP into the process (Section \ref{sec: Fabric means DP}).

\subsection{Description of \Fabric}\label{sec: Fabric means}
We present \Fabric, an ADMM-based framework for solving the balanced clustering problem~\eqref{eq: clustering with st} in VFL settings. 
Fig.~\ref{fig: framework fabricmeans} illustrates the overall framework, which consists of a central server and multiple clients, each holding a different subset of features for the same set of samples. 
The framework operates iteratively through five key steps: 
\ding{192} Each client computes its local objective contribution $Obj_h$ using its local data $\bX_h$ and injects noise; 
\ding{193} Clients upload their local objectives to the server; 
\ding{194} The server aggregates all contributions; 
\ding{195} The server updates the global indicator matrix $\bF$; 
\ding{196} The updated $\bF$ is broadcast to all clients for subsequent optimization until convergence. 
Algorithm~\ref{alg: Fabric-means Combined} provides detailed procedures for both the server and client sides.

\subsubsection{ADMM as a local solver}\label{sec: subproblem Fabric means}
The motivation for adopting ADMM lies in its strong convergence properties and robustness in solving constrained optimization problems. 
To minimize the Lagrangian function over the primal and dual variables, ADMM proceeds by iteratively updating one variable at a time while fixing the others at their most recently updated values. 
To handle the inequality constraints, we further introduce auxiliary variables $\boldsymbol{\theta} = [\theta_1, \ldots, \theta_k]^{\top}$ and reformulate them as equality constraints:
\begin{align}\label{eq:CDKM with ST 2}
\small
\min_{\bF \in \textit{Ind}}\!\! -\!\!\sum_{j=1}^{k} \frac{\bff_j^{\top} \bX^{\top} \bX \bff_j}{\bff_j^{\top} \bff_j}, \, \quad
\text{s.t.} \, \begin{cases}
\theta_{j} = \bff_j^{\top} \bff_j, \\
\alpha_j \leq \theta_{j} \leq \beta_j, 
\end{cases} \!\!\! j \in [k],
\end{align}
where $\theta_j$ represents the size of cluster $j$. 
The corresponding augmented Lagrangian function is:
\begin{equation}\label{eq: balance cluster Lagrangian}
\small
\begin{aligned}
\!\!\!\mathcal{L}(\bF,\! \!\boldsymbol{\theta}, \!\boldsymbol{\lambda})&=\!-\!\sum_{j=1}^{k} \!\!\frac{{\bff}_{j}^{\top}{\bX}^{\top} {\bX}\bff_{j}}{\bff_{j}^{\top} \bff_{j}} 
\!+ \sum_{j=1}^{k}\! \lambda_{j}(\theta_{j} \!- \!{\bff_{j}^{\top} \bff_{j}}) 
+ \!\sum_{j=1}^{k} \!\rho(\theta_{j}\! -\!{\bff_{j}^{\top} \bff_{j}})^2,
\end{aligned}
\end{equation}
where $\boldsymbol{\lambda} \in \mathbb{R}^{k}$ are the Lagrangian multipliers and $\rho > 0$ is the penalty parameter~\cite{Stephen2011Distributed}.

Next, we prove that \eqref{eq: balance cluster Lagrangian} can be decomposed across clients into multiple subproblems that can be solved individually.
Let $\boldsymbol{\lambda_h}$ and $\boldsymbol{\theta_h}$ represent the Lagrangian multipliers and auxiliary variables for the $h$-th client, respectively.
A closer examination reveals that \eqref{eq: balance cluster Lagrangian} is readily decomposable as follows:
\begin{equation}
\small
\begin{aligned}\label{eq: fabric means Lagrangian}
\!\!\overline{\mathcal{L}}(\! \bF,\! \boldsymbol{\theta}, \!\boldsymbol{\lambda})&=\!-\!\sum_{h=1}^{H} \!\sum_{j=1}^{k} \!\! \frac{{\bff}_{j}^{\top}{\bX_h}^{\top}  \! {\bX_h}\bff_{j}}{\bff_{j}^{\top} \bff_{j}} \!+\! \sum_{h=1}^{H}\! \sum_{j=1}^{k}\! \frac{\lambda_{j}}{H} (\theta_{j} \!\!-\!\! {\bff_{j}^{\top} \! \bff_{j}}) \\ 
&+ \sum_{h=1}^{H}\sum_{j=1}^{k} \frac{\rho}{H}(\theta_{j} -{\bff_{j}^{\top} \bff_{j}})^2.
\end{aligned}
\end{equation}
We can present the algorithmic form of ADMM as follows:
\begin{align}\label{eq: admm update order}
\small
\begin{cases}
\boldsymbol{\bF}^{(t)} = \underset{\bF}{\operatorname{argmin}} {\mathcal{L}}(\bF, \boldsymbol{\lambda^{(t-1)} }, \boldsymbol{\theta^{(t-1)}}),\\
\boldsymbol{\theta}^{(t)} = \underset{\boldsymbol{\theta}}{\operatorname{argmin}} \mathcal{L}(\bF^{(t)}, \boldsymbol{\lambda }^{(t-1)}, \boldsymbol{\theta}), \\
\lambda_{j}^{(t)}= \lambda_{j}^{(t-1)}+\rho(\theta_{j}^{(t)} - {(\bff_{j}^{(t)})^{\top} \bff_{j}^{(t)}}),
\end{cases}
\end{align}
where $t$ denotes the iteration step.
Problem \eqref{eq: fabric means Lagrangian} is solved iteratively for $\bF$, $\boldsymbol{\theta}$, and $\boldsymbol{\lambda}$ until convergence.
At each iteration, we update one variable while the others remain fixed at their last values.
For simplicity, we omit the iteration index and denote $\bF$, $\boldsymbol{\lambda}$, and $\boldsymbol{\theta}$ in place of $\bF^{(t)}$, $\boldsymbol{\lambda}^{(t)}$, and $\boldsymbol{\theta}^{(t)}$, respectively.
Next, we provide the detailed derivation for the update of each variable.
\begin{figure}[t]
  \centering
  \includegraphics[width=\linewidth]{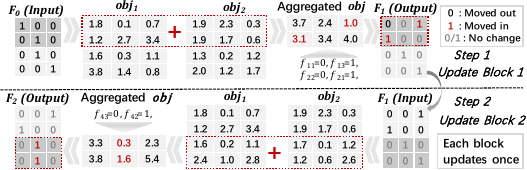}
  \vspace{-1.5em}
  \caption{ An example of \BCD updates $\bF$. The algorithm iteratively updates $\bF$ from $\bF_0$ to $\bF_1$ and then to $\bF_2$. At each step, one block is updated while the others are fixed.
   }
  \vspace{-1em}
  \label{fig: updatF}
\end{figure}

\myparagraph{$\bF$-solving procedure} 
In VFL, updating each row of $\bF$ individually requires one communication round per sample, incurring substantial overhead.
To reduce communication rounds, we use \BCD~\cite{Zhu23F3KM}, where the client communicates with the server only after computing $n_b$ rows of the objective, with $n_b$ being the block size. 
\BCD updates a block of $n_b$ samples simultaneously in each iteration, significantly reducing communication frequency in VFL.

Fig.~\ref{fig: updatF} illustrates the block-coordinate update procedure for the indicator matrix $\bF$. 
Starting from $\bF_0$, the algorithm updates one block of $\bF$ at a time while keeping the remaining blocks fixed. 
For the selected block, each client computes its local objective values ($obj_1$ and $obj_2$), which are then aggregated to determine the best cluster assignments. 
The corresponding rows of $\bF$ are updated in a batch manner: an entry is set to 1 if the sample is assigned to the cluster and to 0 otherwise. 
Repeating this procedure generates a sequence of assignment matrices, $\bF_1, \bF_2, \ldots$, until convergence, thereby improving the efficiency of the overall optimization.

To reduce the complexity of the objective function and facilitate the subsequent derivation, we introduce a set of auxiliary matrices $\{\bF_i^{(0)}, \bF_i^{(1)}, \dots, \bF_i^{(k)}\}$. 
Here, $\bF_i^{(m)}$ denotes the candidate assignment matrix associated with the $i$-th sample, and $\bF_i^{(0)}$ is defined as the same as $\bF_i^{(m)}$ except that its $i$-th row is set to the zero vector. 
With this notation, Problem~\eqref{eq: fabric means Lagrangian} can be equivalently reformulated as follows:
\begin{equation}
\small
\begin{aligned} \label{eq:fabric means Lyapunov}
&\overline{\mathcal{L}}(m, i, h) = \mathcal{L}(\bF_{i}^{(m)}, \boldsymbol{\lambda}, \boldsymbol{\theta}) 
- \mathcal{L}(\bF_{i}^{(0)}, \boldsymbol{\lambda}, \boldsymbol{\theta})  \\
&= - \frac{(\bff_{m}^{(m)})^{\top} \bX_h^{\top} \bX_h \bff_{m}^{(m)}}{(\bff_{m}^{(m)})^{\top} \bff_{m}^{(m)}} + 
\frac{(\bff_{m}^{(0)})^{\top} \bX_h^{\top} \bX_h \bff_{m}^{(0)}}{(\bff_{m}^{(0)})^{\top} \bff_{m}^{(0)}} \\
& + \frac{\lambda_{m}}{H} \Big(\big( (\bff_{m}^{(0)})^{\top} \bff_{m}^{(0)} - 
(\bff_{m}^{(m)})^{\top} \bff_{m}^{(m)} \big) + \rho \big( (\bff_{m}^{(m)})^{\top} \bff_{m}^{(m)} \big)^2 \Big)  \\
& - \frac{2\theta_{m}}{H} \Big(\big((\bff_{m}^{(m)})^{\top} \bff_{m}^{(m)}-
(\bff_{m}^{(0)})^{\top} \bff_{m}^{(0)} \big) - \rho \big( (\bff_{m}^{(0)})^{\top} \bff_{m}^{(0)} \big)^2\Big).
\end{aligned}
\end{equation}

Let $\bF$ denote the current indicator matrix preloaded into memory. 
Let $p \in [k]$ denote the current cluster assignment of the sample such that the $p$-th element of the $i$-th row of $\bF$ is 1.
Define $\bff_{m}^{(m)}$ as the $m$-th column of the matrix $\bF_i^{(m)}$, $\bff_{m}^{(0)}$ as the $m$-th column of $\bF_i^{(0)}$, and $\bff_m$ as the $m$-th column of $\bF$.
According to the conditions specified in \eqref{eq:fabric means Lyapunov}, the following two cases are considered:
\begin{itemize}
\item \textbf{Situation 1:}
When $m=p$, the $i$-th entries of both $\bff_{m}^{(m)}$ and $\bff_m$ are equal to 1, which implies that $\bff_{m}^{(m)}=\bff_m$.  
Hence, $\bff_m^{(m)}=\bff_m$.
Define $\boldsymbol{\gamma}_m=\bff_m-\bff_m^{(0)}$, whose $i$-th entry is 1 and all other entries are 0.  
Thus, we obtain
\begin{align*}
\small
\bX_h \bff_m = \bX_h \bff_m^{(0)} + \bx_i^h, \quad (\bff_m^{(0)})^\top \bff_m^{(0)} = \bff_m^\top \bff_m - 1.
\end{align*}

Substituting the above equations into \eqref{eq:fabric means Lyapunov}, we obtain:
\begin{align}\label{eq:fabric means Lyapunov1}
\small
&\overline{\mathcal{L}}(m, i, h) =\!\! -\!\! \frac{\bff_{m}^{\top} \bX_h^{\top} \bX_h \bff_{m}}{\bff_m^{\top} \bff_{m}}  + \frac{\rho}{H} ( 2 \bff_{m}^{\top} \bff_{m} \!-\! 2 \theta_{m}  \!- \!1 )  \nonumber \\
& \!\!+ \!\! \frac{\bff_{m}^{\top} \bX_h^{\top}\!\! \bX_h \bff_m \!\!-\! \!2 (\bx_i^{h})^{\top} \bX_h \bff_m \!\!+\! (\bx_i^{h})^{\top} \bx_i^{h}}{\bff_{m}^{\top} \bff_{m} - 1}\!\! + \!\!\frac{\lambda_m}{H}.
\end{align}

\item \textbf{Situation 2:}
When $ m \neq p $, the $i$-th element of both $\bff_{m}^{(m)}$ and $\bff_{m}$ is 0, which implies that $\bff_{m}^{(0)}$ = $\bff_{m}$.
We introduce $\boldsymbol{\gamma}_{m}=\bff_{m}^{m} -\bff_{m}$ where the $i$-th element is 1 and all other elements are 0.
Thus, we can derive the following:
\begin{align*}
\small
\begin{array}{c}
\bX_h \bff_{m}^{(m)}=\bX_h(\bff_{m}+\boldsymbol{\gamma}_{m})=\bX_h \bff_{m}+\bx_{i}^h, \\ 
(\bff_{m}^{(m)})^{\top}\bff_{m}^{(m)}=(\bff_{m}+\boldsymbol{\gamma}_{m})^{\top}(\bff_{m} + \boldsymbol{ \gamma}_{m} ) =\bff_{m}^{\top} \bff_{m} + 1.\end{array}
\end{align*}
Substituting the above equations into \eqref{eq:fabric means Lyapunov}, we obtain:
\begin{align}\label{eq:fabric means Lyapunov2}
\small
& \overline{\mathcal{L}}(m, i, h) =\frac{\bff_{m}^{\top} \bX_h^{\top} \bX_h \bff_{m}}{\bff_{m}^{\top} \bff_{m}}  +  \frac{\rho}{H} (2\bff_{m}^{\top}\bff_{m} - 2 \theta_{m} + 1  ) \nonumber \\
&\!\!\!-\!\frac{\bff_m^{\top} \bX_h^{\top} \bX_h \bff_m \!\!-\! 2(\bx_i^{h})^{\top}\! \bX_h \bff_m\!\! + \!(\bx_i^{h})^{\top} \bx_i^{h}}{\bff_m^{\top} \bff_{m} \!\!+\! \!1 }  \!\!+\!\! \frac{\lambda_m}{H}.
\end{align}
\end{itemize}
The server aggregates local objectives from all clients and assigns sample $i$ to the cluster, minimizing the total objective:
\begin{equation}
\label{eq:abmeans_UpdateF}
\small
f_{iq} =
\begin{cases}
1, & q = \arg \min_{m}  \sum_{h=1}^{H}\overline{\mathcal{L}}(m, i, h), \\
0, & \text{otherwise.}
\end{cases}
\end{equation}
To maintain non-empty clusters during the update, if sample $i$ currently belongs to cluster $r$ and $\bff_r^{\top}\bff_r=1$, the server rejects all candidate moves with $m\neq r$ and keeps sample $i$ in cluster $r$. 
Equivalently, these infeasible candidate moves are assigned an infinite objective value before taking the minimum in \eqref{eq:abmeans_UpdateF}.

After updating row $i$ from cluster $r$ to cluster $q$, we incrementally update cached statistics~\cite{nie2022CDKM}:
\begin{align}
\small
&\hspace{2.5em} \bff_r^{\top} \bff_r \gets \bff_r^{\top} \bff_r - 1,  
\bff_q^{\top} \bff_q \gets \bff_q^{\top} \bff_q + 1,
\label{eq: Update_F} \\
&\hspace{2.0em} \bX_h \bff_r \gets \bX_h \bff_r - \bx_i^{h}, 
\bX_h \bff_q \gets \bX_h \bff_q + \bx_i^{h}, \notag \\
&\bff_r^{\top} \bX_h^{\top} \bX_h \bff_r 
\gets \bff_r^{\top} \bX_h^{\top} \bX_h \bff_r 
- 2 (\bx_i^{h})^{\top} \bX_h \bff_r 
+ (\bx_i^{h})^{\top} \bx_i^{h}, \notag \\
&\bff_q^{\top} \bX_h^{\top} \bX_h \bff_q 
\gets \bff_q^{\top} \bX_h^{\top} \bX_h \bff_q 
+ 2 (\bx_i^{h})^{\top} \bX_h \bff_q 
+ (\bx_i^{h})^{\top} \bx_i^{h}. \notag
\end{align}
\begingroup
\begin{algorithm}[t]
    \caption{\Fabric}
    \label{alg: Fabric-means Combined}
    \renewcommand{\algorithmicrequire}{\textbf{Input:}}
    \renewcommand{\algorithmicensure}{\textbf{Output:}}

    \KwIn{$T$: outer iterations, $H$: number of clients, local datasets $\bX_h$, number of clusters $k$, block size $n_b$, clipping threshold $\Delta$, target privacy budget $(\epsilon,\delta)$, and DP flag \texttt{useDP}.}
    \KwOut{ Cluster indicator matrix $\bF$.}

    \textcolor{gray}{\underline{\textit{Server initializes:}}}  $\bF$, $\boldsymbol{\theta}$, and $\boldsymbol{\lambda}$;
    
    Divide data indices into $n/n_b$ blocks;
    
    \textcolor{gray}{\underline{\textit{Clients initialize:}}}$\bX_h \bff_m$, $\bff_m^{\top} \bff_m$, $\bff_m^{\top} \bX_h^{\top} \bX_h \bff_m$, $(\bx_j^{h})^{\top} \bx_j^{h}$, $m \in [k], j \in [n]$;
        
    \For{$t=0,1,\cdots,T-1$}{

        \For{each block $l \in [n/n_b]$}{

            \For{each client $h = 1, \dots, H$}{
            \textcolor{gray}{\underline{\textit{Calculate Obj:}}} Compute $\overline{\mathcal{L}} (m, j, h)$,  $m \in [k]$, $j \in \text{block } l$ as follows:  \\
            \uIf{$use DP$}{
                Using Eq.~\eqref{eq:fabric means DP Lagrangian};
              }
              \uElse{
                Using Eq.~\eqref{eq:fabric means Lyapunov1} and Eq.~\eqref{eq:fabric means Lyapunov2};
              }            
            }
                
            \textcolor{gray}{\underline{\textit{Aggregate:}}} Aggregate $Obj$ from all clients;
            
            \textcolor{gray}{\underline{\textit{Server Updates:}}} Update $\bF^t$ , $\boldsymbol{\theta^t}$ and $\boldsymbol{\lambda^t}$ via Eq.~\eqref{eq:abmeans_UpdateF}, Eq.~\eqref{eq:Updatetheat} and Eq.~\eqref{eq:Updalambda};
            
            \For{each client $h = 1, \dots, H$}{
                 \textcolor{gray}{\underline{\textit{Download:}}} Download $\bF$, $\boldsymbol{\theta}$, $\boldsymbol{\lambda}$;
                 
                \textcolor{gray}{\underline{\textit{Local update:}}} Update $\bff_m^{\top} \bX_h^{\top} \bX_h \bff_m$, $\bff_m^{\top} \bff_m$, and $\bX_h \bff_m$ using Eq.~\eqref{eq: Update_F};
            }
        }
    }
    \KwRet  $\bF$.
\end{algorithm}
\endgroup 

\myparagraph{$\boldsymbol{\theta}$ and $\boldsymbol{\lambda}$-solving procedure}
When solving for $\boldsymbol{\theta}$, we treat $\bF$ and $\boldsymbol{\lambda}$ as fixed. 
The optimization in \eqref{eq: fabric means Lagrangian} becomes a quadratic program with respect to $\boldsymbol{\theta}$:
\begin{align}\label{eq:Update theat 2}
\small
\begin{aligned}
\quad \boldsymbol{\theta}_{j} = &  \underset{\theta}{\operatorname{argmin}} \rho\  \left(  \theta_j - \bff_{j}^{\top} \bff_{j}  \right)^2 + \lambda_j \left(  \theta_j - \bff_{j}^{\top} \bff_{j}  \right) \\
= & \underset{\theta}{\operatorname{argmin}} \rho\ {\theta_j}^2 - \left(2 \rho \bff_{j}^{\top} \bff_{j}  + \lambda_j \right)  \theta_j + \rho \left(\bff_{j}^{\top} \bff_{j}  \right)^2.
\end{aligned}
\end{align}
Therefore, the subproblem with respect to $\boldsymbol{\theta}$ admits a closed-form solution, and the corresponding update is given by
\begin{align}\label{eq:Updatetheat}
\small
\begin{aligned}
\theta_{j} = 
\begin{cases} 
\bff_{j}^{\top} \bff_{j} - \frac{\lambda_{j}}{2\rho}, & \alpha_j \leq \theta_{j} \leq \beta_j, \\
\beta_j, & \beta_j < \theta_{j}, j \in [k],\\
\alpha_j, & \theta_{j}  < \alpha_j.
\end{cases}
\end{aligned}
\end{align}

To compute the dual variable $\boldsymbol{\lambda}$, we leverage the intrinsic features of the ADMM framework \cite{Stephen2011Distributed}, enabling the derivation of an update formula for $\boldsymbol{\lambda}$:
\begin{align}\label{eq:Updalambda}
\small
\begin{aligned}
{\lambda}_{j} = &{\lambda}_{j} + \rho (\theta_j - \bff_{j}^{\top} \bff_{j}  ), \,\quad j \in [k].
\end{aligned}
\end{align}
Compared to updating one row at a time, \BCD updates $\bF$ less frequently per iteration, which may decrease its accuracy. 
Therefore, it is crucial to carefully choose the value of $n_b$ to balance accuracy and computational efficiency.

\subsubsection{Adapting DP}\label{sec: Fabric means DP}
To protect client data from membership inference attacks while preserving clustering utility, we integrate DP into \Fabric by injecting calibrated Gaussian noise into the objective function computation at each client. 
This ensures that the objective messages sent from clients to the server are derived from DP-protected statistics rather than raw data.
Specifically, in the $\bF$-solving procedure, each client computes its local objective value $\overline{\mathcal{L}}(m,i,h)$ for assigning sample $i$ to cluster $m$. Before uploading this value to the server, the client first clips it into a bounded interval and then adds Gaussian noise. 
Specifically, the objective function with Gaussian noise is defined as
\begin{align}\label{eq:fabric means DP Lagrangian}
\small
\!\!\!\!\!\!\mathcal{M}(m, i, h) \!= \!\max\!\Big(\!\min\big(\overline{\mathcal{L}}(m, i, h), \Delta \big),\! \!-\!\Delta \Big) \!+\! \mathcal{N}(0, \sigma^2),
\end{align}
where $\Delta>0$ is the clipping threshold that bounds the objective contribution, and $\mathcal{N}(0, \sigma^2)$ represents Gaussian noise with variance $\sigma^2$. 
To make the privacy accounting explicit, we use the Renyi differential privacy (RDP) accountant, equivalently the moments accountant for Gaussian mechanisms~\cite{mironov2017renyi,abadi2016deep}. 
We state the guarantee from the perspective of each client's messages to the server.
Let $Q_h$ be the number of noisy objective releases made by client $h$ during one execution of \Fabric.
In our implementation, when each candidate objective is treated as a single scalar query, $Q_h$ is upper-bounded by $Tnk$.
More generally, $Q_h$ can be replaced by the actual number of clipped objective queries released by client $h$.
For a Gaussian mechanism with sensitivity upper bound $\Gamma$ and noise standard deviation $\sigma$, the RDP privacy loss at order $\alpha>1$ after composing all $Q_h$ scalar releases is
\begin{equation}\label{eq:privacy accountant}
\small
\epsilon_{\mathrm{RDP},h}(\alpha)
=
Q_h\cdot \frac{\alpha \Gamma^2}{2\sigma^2}.
\end{equation}
Converting RDP to approximate DP gives the following privacy accountant:
\begin{equation}\label{eq:rdp dp conversion}
\small
\epsilon_h(\sigma,\delta)
=
\min_{\alpha \in \mathcal{A},\,\alpha>1}
\left(
Q_h\cdot \frac{\alpha \Gamma^2}{2\sigma^2}
+\frac{\ln(1/\delta)}{\alpha-1}
\right),
\end{equation}
where $\mathcal{A}$ is a finite set of candidate Renyi orders. 
For scalar clipping to $[-\Delta,\Delta]$, we set $\Gamma=2\Delta$ unless a tighter valid sensitivity upper bound is available.
Given a target budget $(\epsilon,\delta)$, we choose the smallest $\sigma$ such that $\epsilon_h(\sigma,\delta)\le \epsilon$ for each client, which can be solved by a one-dimensional binary search before running the algorithm.

As shown in Algorithm~\ref{alg: Fabric-means Combined}, when \texttt{useDP} is enabled, each client uploads the noisy objective value in \eqref{eq:fabric means DP Lagrangian} instead of the exact objective values in \eqref{eq:fabric means Lyapunov1} and \eqref{eq:fabric means Lyapunov2}. 
The server then updates $\bF$, $\boldsymbol{\theta}$, and $\boldsymbol{\lambda}$ using only these perturbed values. 
When \texttt{useDP} is disabled, each client directly uploads the exact objective values, and the server performs the same updates without noise injection. 
Since the privacy-preserving mechanism requires only clipping and one Gaussian sample per objective evaluation, it introduces negligible computational overhead in practice while retaining useful clustering performance.

In \Fabric, the target privacy budget $\epsilon$ determines the overall trade-off between privacy protection and clustering utility through the per-client accountant in \eqref{eq:rdp dp conversion}. Specifically, a smaller $\epsilon$ requires a larger noise scale $\sigma$, which provides stronger privacy protection but may reduce clustering utility. In contrast, a larger $\epsilon$ permits less perturbation, which generally leads to better clustering performance, although at the cost of weaker privacy guarantees.
We now present the standard per-client message-level privacy guarantee under this accounting.



\subsection{Privacy Analysis}
We now analyze the privacy guarantees of the proposed perturbation mechanism in \Fabric. 
In our framework, the server aggregates noisy local objectives from clients across all block updates and optimization iterations. 
We first establish the RDP guarantee for a single scalar release, and then analyze the per-client privacy loss under the accountant in \eqref{eq:rdp dp conversion}. 
Recall that at iteration $t$, for each client $h$, data sample $i$, and candidate cluster $m$, the mechanism evaluates the local objective value $\overline{\mathcal{L}}(m, i, h)$. 
To bound the sensitivity of this objective value, we first clip it to a maximum absolute value of $\Delta$. Let 
\begin{equation}
\small
g(m, i, h)
=
\max\Big(\min\big(\overline{\mathcal{L}}(m,i,h),\Delta\big),-\Delta\Big).
\end{equation}
denote the clipped objective evaluated on the local dataset $X_h$ at iteration $t$. 
Here, for notational simplicity, we keep the dependence of $\overline{\mathcal{L}}(m,i,h)$ and $g(m,i,h)$ on the local dataset $X_h$ implicit. 

\begin{theorem}\label{thm:dp_guarantee_single} 
For any single iteration $t$ and fixed indices $(m,i,h)$, assuming the $\ell_2$-sensitivity of the clipped objective value $g(m,i,h)$ is bounded by $\Gamma$ under the neighboring relation in Definition~\ref{def:dp}, the local Gaussian mechanism $\mathcal{M}$ satisfies $(\alpha,\epsilon_{\mathrm{RDP}}^{(1)}(\alpha))$-RDP for any $\alpha>1$, where
\begin{equation}
\small
\epsilon_{\mathrm{RDP}}^{(1)}(\alpha)
=
\frac{\alpha \Gamma^2}{2\sigma^2}.
\end{equation}
\end{theorem}

\begin{proof} For a fixed iteration $t$ and fixed indices $(m,i,h)$, the released quantity is the scalar query $g(m,i,h)$ evaluated on the local dataset $X_h$.
By clipping, the released scalar is bounded in the range $[-\Delta,\Delta]$, so its global sensitivity is at most $2\Delta$ under the neighboring relation. Thus, $\Gamma=2\Delta$ is a valid conservative upper bound.
That is, for any two neighboring datasets $X_h$ and $X_h'$, we have
\begin{equation}
\small
\left| g(m,i,h;X_h) - g(m,i,h;X_h') \right| \le \Gamma .
\end{equation}
Since $\mathcal{M}(X)$ perturbs a scalar objective value with sensitivity at most $\Gamma$ using additive Gaussian noise, it directly instantiates the Gaussian mechanism. 
The Gaussian mechanism with $\ell_2$-sensitivity $\Gamma$ satisfies $(\alpha,\alpha \Gamma^2/(2\sigma^2))$-RDP for any $\alpha>1$~\cite{mironov2017renyi}. 
\end{proof} 

\begin{theorem}\label{thm:dp_guarantee_total}
For each client $h$, under the honest-but-curious server model and the bounded clipped-query assumption above, the sequence of messages sent from client $h$ to the server satisfies $(\epsilon_h,\delta)$-DP with respect to client $h$'s local data, where $\epsilon_h$ is computed by the per-client RDP accountant in \eqref{eq:rdp dp conversion}. In particular, if $\sigma$ is chosen such that $\epsilon_h\le \epsilon$, then the released messages satisfy $(\epsilon,\delta)$-DP.
\end{theorem}

\begin{proof}
The releases in \Fabric are adaptive: the global variables $\boldsymbol{F}$, $\boldsymbol{\theta}$, and $\boldsymbol{\lambda}$ at a later step depend on the noisy outputs of previous steps. 
By Theorem~\ref{thm:dp_guarantee_single}, each scalar release from client $h$ is $(\alpha,\alpha \Gamma^2/(2\sigma^2))$-RDP. 
RDP composes additively, even under adaptive composition~\cite{mironov2017renyi}. 
Thus, after client $h$ releases at most $Q_h$ scalars, the client-to-server transcript is
\begin{equation}
\small
\left(
\alpha,\,
Q_h\cdot \frac{\alpha \Gamma^2}{2\sigma^2}
\right)\text{-RDP}
\end{equation}
for any $\alpha>1$. 
The standard conversion from RDP to approximate DP implies that this message transcript is $(\epsilon_h(\sigma,\delta),\delta)$-DP, where $\epsilon_h(\sigma,\delta)$ is given by \eqref{eq:rdp dp conversion}. 
Choosing $\sigma$ such that $\epsilon_h(\sigma,\delta)\le \epsilon$ yields $(\epsilon,\delta)$-DP for client $h$'s released messages. 
Finally, the updates of $\bF$, $\boldsymbol{\theta}$, and $\boldsymbol{\lambda}$ are deterministic post-processing of the noisy objective values and therefore incur no additional privacy loss.
\end{proof}

\subsection{Discussions} 
\label{sec: Discussions} 
This section discusses the privacy advantages of \Fabric in VFL, and its robustness to data heterogeneity across parties.

\myparagraph{Privacy preservation compared with \DPFMPS} 
Unlike existing methods such as \DPFMPS that transmit DP-protected cluster centers and membership encodings, \Fabric exchanges only scalar objective values and a binary assignment matrix $\bF$. 
By avoiding center transmission, our method prevents adversaries from reconstructing cluster geometries, thereby providing stronger privacy protection.

\myparagraph{Robustness to heterogeneity} 
In VFL, clients often hold vastly different numbers of features or semantically diverse feature types, which can challenge clustering algorithms. 
\Fabric naturally handles this heterogeneity through its decomposition structure.
The global objective term $\bff_j^{\top} \bX^{\top} \bX \bff_j$ is the exact sum of locally computed terms $\sum_{h=1}^{H} \bff_j^{\top} \bX_h^{\top} \bX_h \bff_j$, making the outcome theoretically invariant to how features are partitioned. 
This robustness persists even in the presence of DP noise.
Although noise is introduced into the server-side aggregation at each round, its effect does not accumulate without bound.
At each iteration, clients update their local computations based on the latest global $\bF$ returned by the server, which continually realigns the optimization process.
Such iterative resynchronization helps maintain a stable optimization trajectory, resulting in nearly identical performance across diverse feature distributions.

\subsection{Complexity Analysis} 
In this part, we analyze the communication and computational complexity of \Fabric.

\subsubsection{Communication complexity}
Each iteration of Algorithm~\ref{alg: Fabric-means Combined} consists of $b = n/n_b$ rounds for updating $\bF$ in blocks and one round for synchronizing $\boldsymbol{\lambda}$ and $\boldsymbol{\theta}$. 
Over $T$ iterations, the total communication complexity is $O(bT)$, which can be controlled by adjusting the block size $n_b$.

\subsubsection{Computational complexity}
Next, we analyze the computational complexity, focusing on multiplication and division operations, which dominate the overall computational cost.

\myparagraph{Client-side} 
Each client first precomputes $\bX_h \bff_j$ for all $k$ clusters and $\bff_j^{\top} \bff_j$ for normalization, incurring a one-time cost of $O(nd_hk)$.  
During each iteration, for every block $m$, the client evaluates $\overline{\mathcal{L}}(m, j, h)$ across all clusters. 
With the precomputed statistics, this requires $O(n_bd_hk)$ operations per block, resulting in a total cost of $O(nd_hk)$ over all $b$ blocks in one iteration. 
After assignment, updating the corresponding precomputed statistics costs $O(kd_h)$ per block, contributing an additional $O(nkd_h/n_b)$ per iteration. 
Thus, the total complexity on each client side over $T$ iterations is $O(Tnd_hk)$.

\begin{table}[ht]
\centering
\renewcommand{\arraystretch}{1.1}
\setlength{\tabcolsep}{5.5pt} 
\caption{An overview of the datasets.}
\label{table:Dataset Table}
  \vspace{-0.5em}
\resizebox{0.8\columnwidth}{!}{
\begin{tabular}{clcc}
\hline
No. & Datasets                & Instances & Features \\ \hline
1 & \Census \cite{Bera19Fair} & 32,561   & 15       \\
2 & \Diabetes \cite{burrows2017incidence} & 4,747      & 21       \\
3 & \Letter \cite{slateletter}   & 1,339      & 7        \\
4 & \Loan \cite{2017loan}  & 60,000    & 120       \\
5 & \Patient \cite{strack2014impact} & 101,766     & 50      \\
6 & \Student \cite{Kuzilek2015open} & 32,594     & 12 \\
7 & \Taxi \cite{meg2017taxi}    & 2,509      & 34       \\
8 & \YearPredictionMSD \cite{2017year_prediction_msd} & 515,345       & 90      \\
9 & \seconddata \cite{Bera19Fair} & 2,458,285   & 69       \\
10 & \HMDA \cite{HMDA2017} & 5,986,660   & 53       \\ \hline
\end{tabular}
}
\vspace{-1.4em}
\end{table}

\myparagraph{Server-side}
In each iteration, the server aggregates $nk$ objective values from clients, incurring a computational cost of $O(nk)$. 
Updating $\bF$ involves selecting optimal assignments for $n$ samples across $k$ clusters, costing $O(nk)$. 
Parameter updates for $\boldsymbol{\theta}$ and $\boldsymbol{\lambda}$ require $O(k)$ operations. 
Thus, the total complexity over $T$ iterations is $O(Tnk)$. 
Both client and server complexities scale linearly in $n$, $d$, and $k$, ensuring efficiency for practical VFL deployments.

\begin{figure*}[t]
  \centering
  \includegraphics[width=\linewidth]{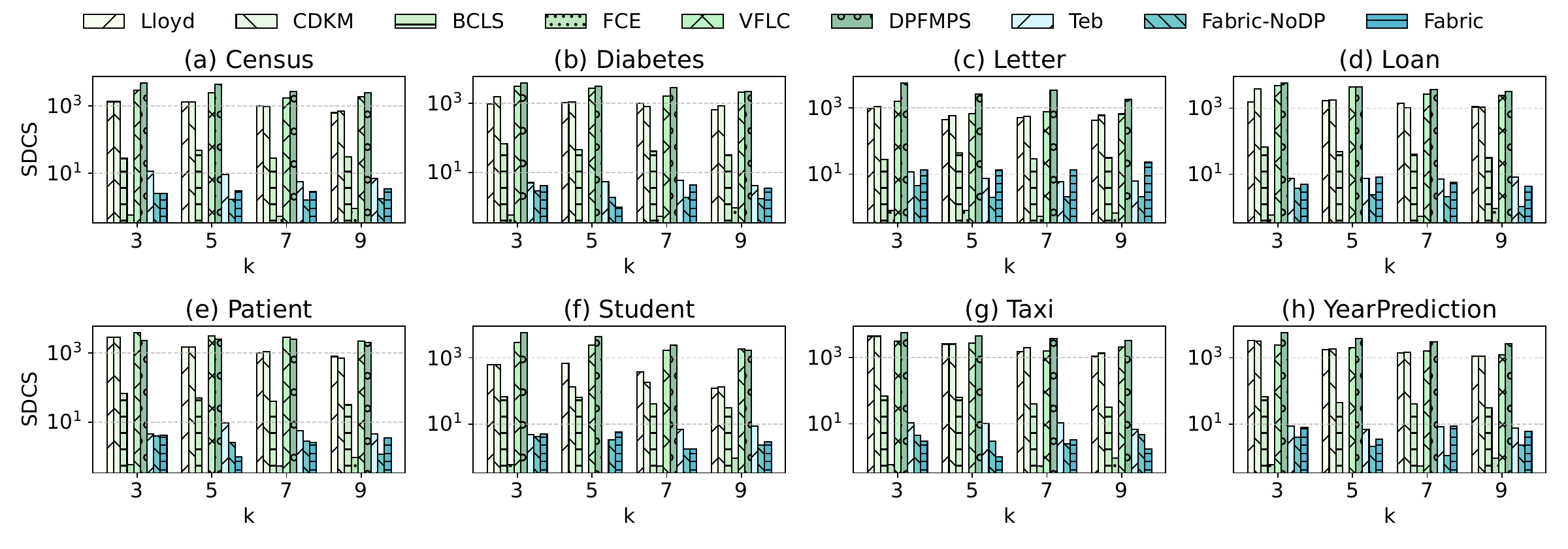}
  \vspace{-2.1em}
  \caption{Balance performance ($SDCS$) of different methods on different datasets at varying $k$.}
  \vspace{-1.4em}
  \label{fig: Balance1}
\end{figure*}
\begin{figure*}[t]
    \centering
    \includegraphics[width=\linewidth]{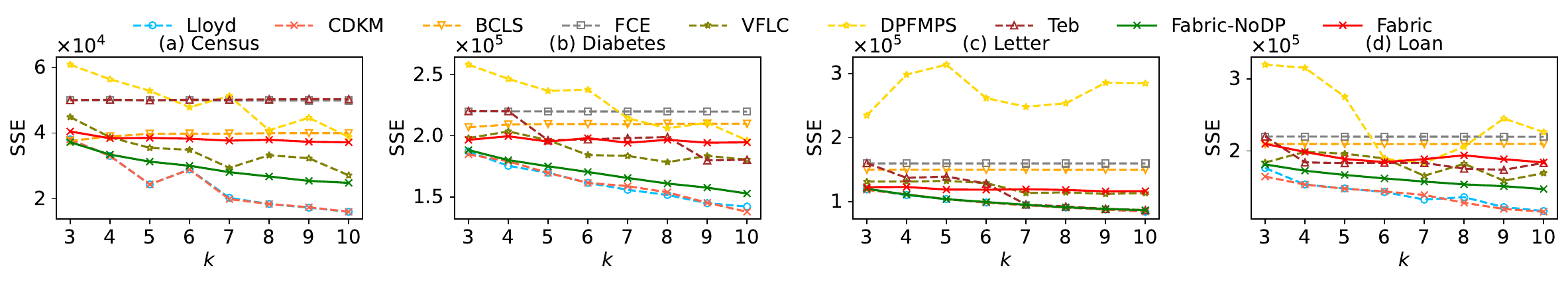}
  \vspace{-2.6em}
\end{figure*}
\begin{figure*}[t]
    \centering
    \includegraphics[width=\linewidth]{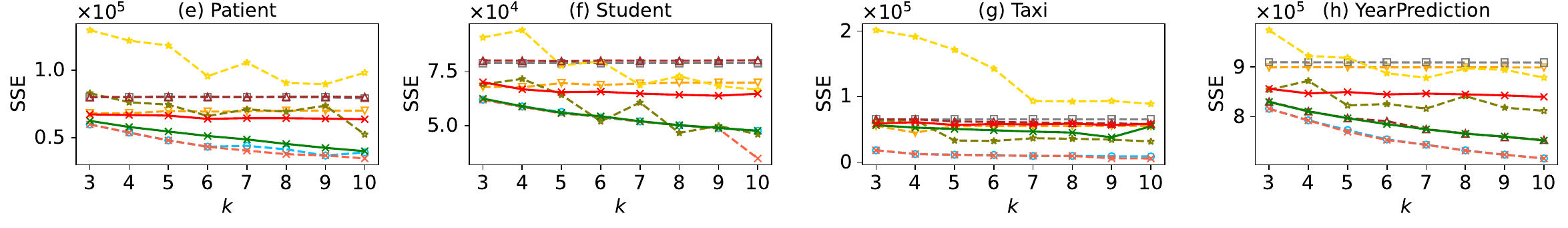}
  \vspace{-2.6em}
    \caption{$SSE$ of different methods on different datasets at varying $k$.}
  \vspace{-1.5em}
    \label{fig: SSE1 at varying k}
\end{figure*}
\begin{figure*}[t]
  \centering
  \includegraphics[width=\linewidth]{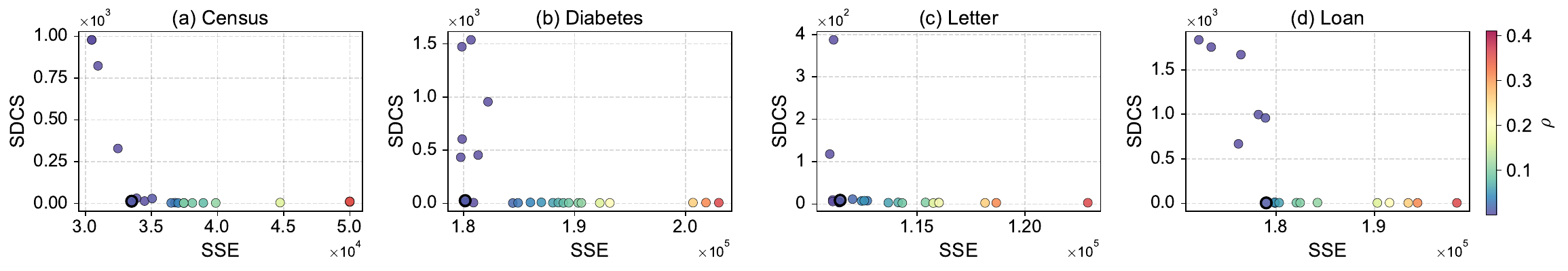}
  \vspace{-1.8em}
  \caption{$SDCS$ and $SSE$ of \Fabric on different datasets  at varying $\rho$.}
   \vspace{-2.em}
  \label{fig:pareto_rho}
\end{figure*}
\begin{figure*}[t]
    \centering
    \includegraphics[width=\linewidth]{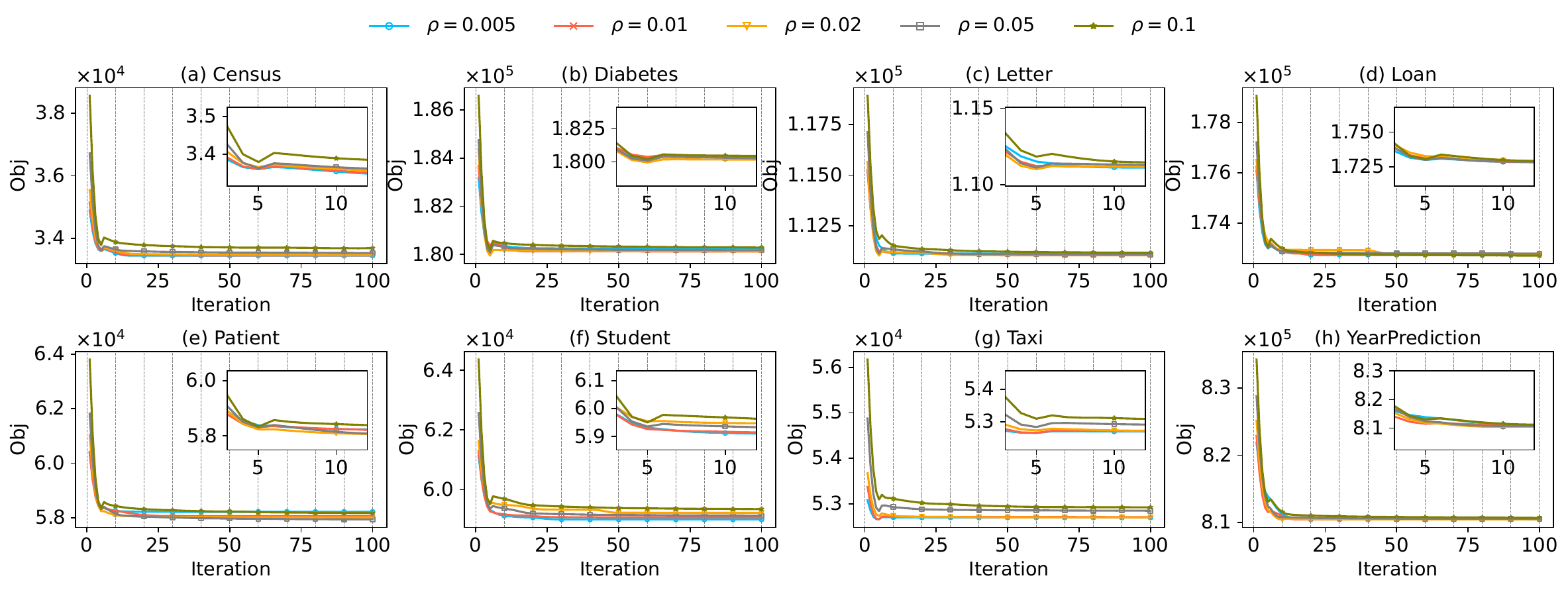}
   \vspace{-2em}
    \caption{The convergence of \FabricNoDP  in terms of varying $\rho$.}
  \vspace{-1.4em}
    \label{fig: Fabric-means SSE at varying rho}
\end{figure*}
\begin{figure*}[t]
    \centering
    \includegraphics[width=\linewidth]{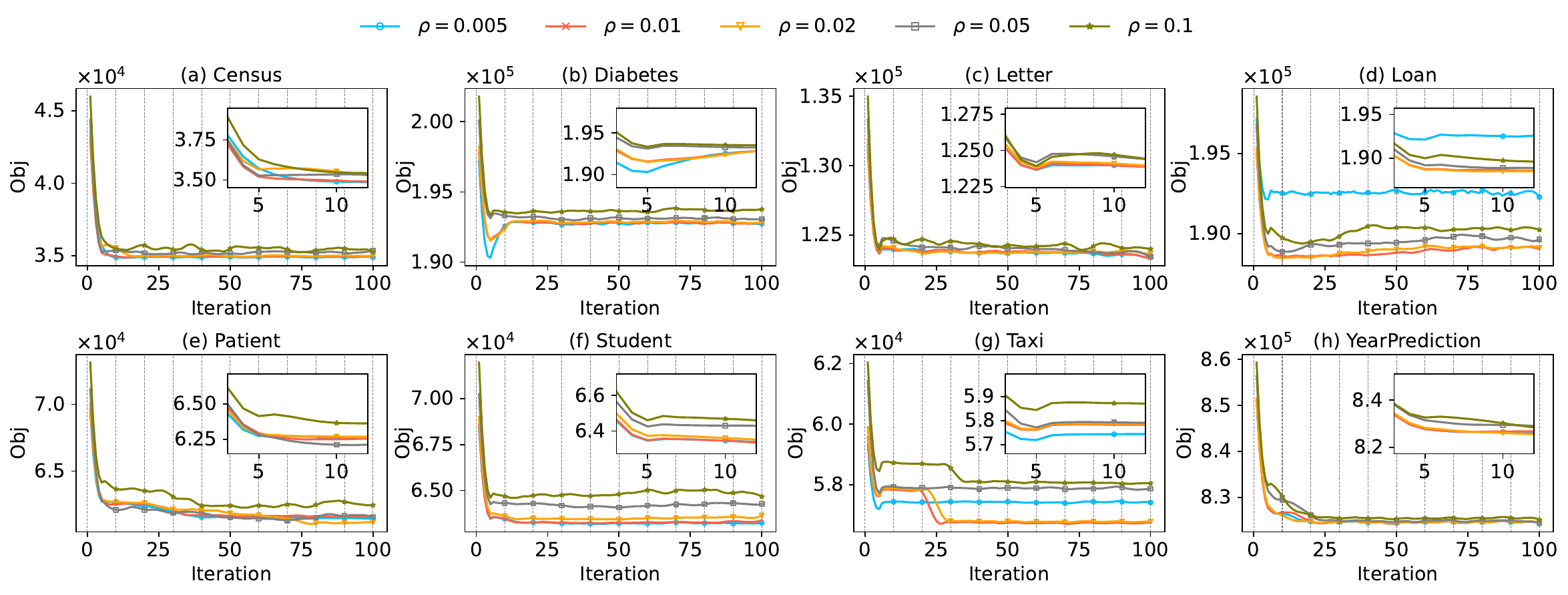}
   \vspace{-2em}
    \caption{The convergence of \Fabric  in terms of varying $\rho$.}
  \vspace{-1.4em}
    \label{fig: Fabric-meansdp SSE1 at varying rho}
\end{figure*}
\begin{figure*}[t]
  \centering
  \includegraphics[width=\linewidth]{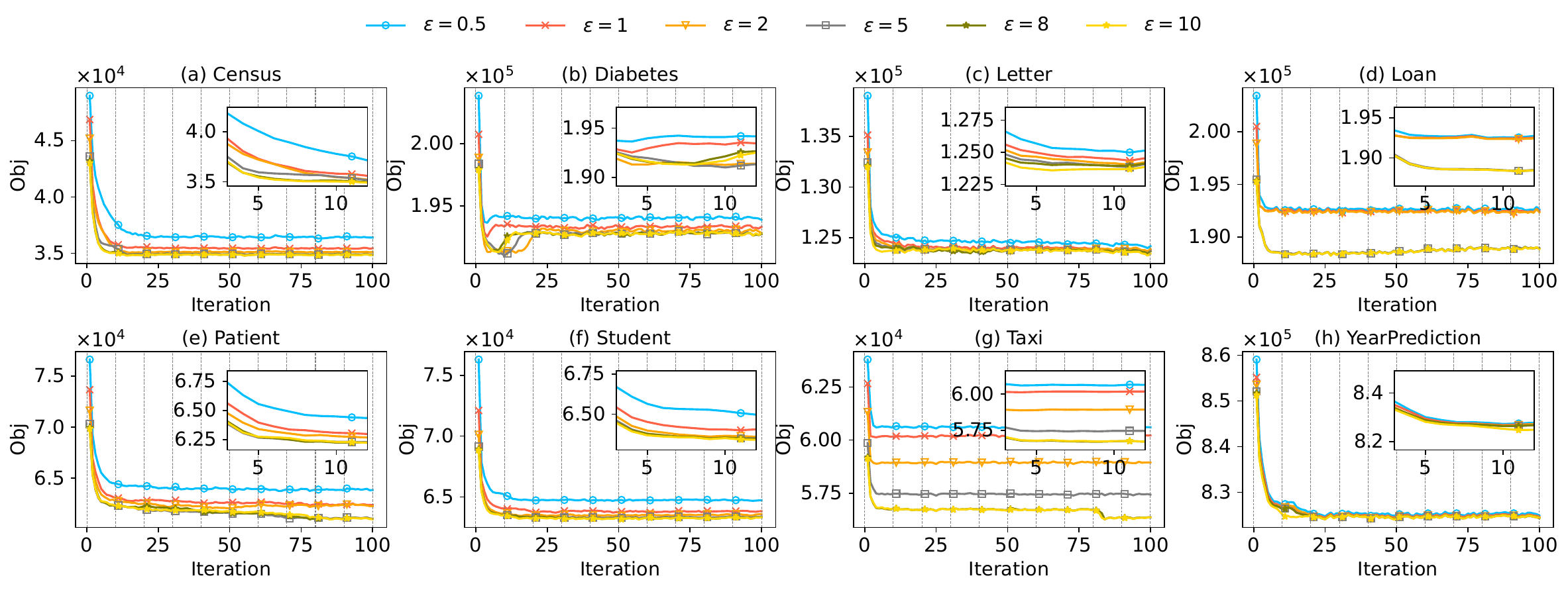}
  \vspace{-2em}
    \caption{The objective value of \Fabric under different privacy budgets $\epsilon$.}
  \vspace{-1.5em}
    \label{fig: Fabric-means-dp SSE1 at varying epsilon}
\end{figure*}

\begin{figure*}[t]
  \centering
    \includegraphics[width=\linewidth]{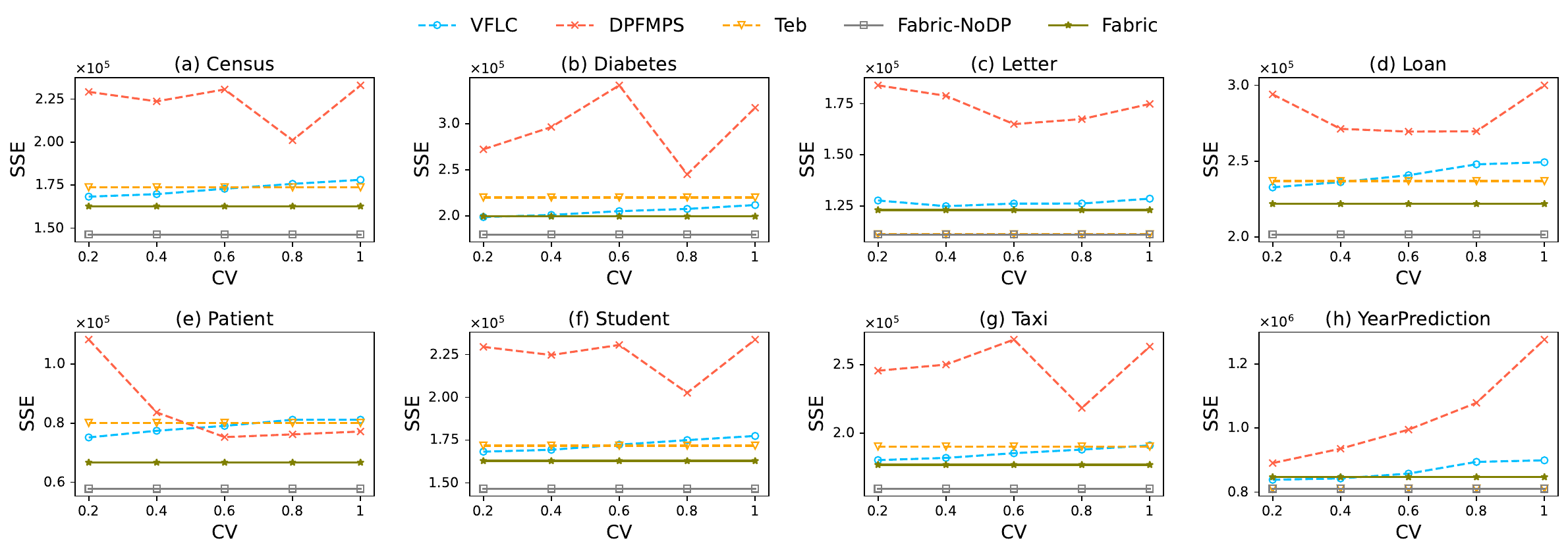}
  \vspace{-2em}
    \captionof{figure}{The $SSE$ at varying CV.}
      \vspace{-2em}
    \label{fig:  SSE at varying CV}
\end{figure*}
\begin{figure*}[t]
  \centering
    \includegraphics[width=\linewidth]{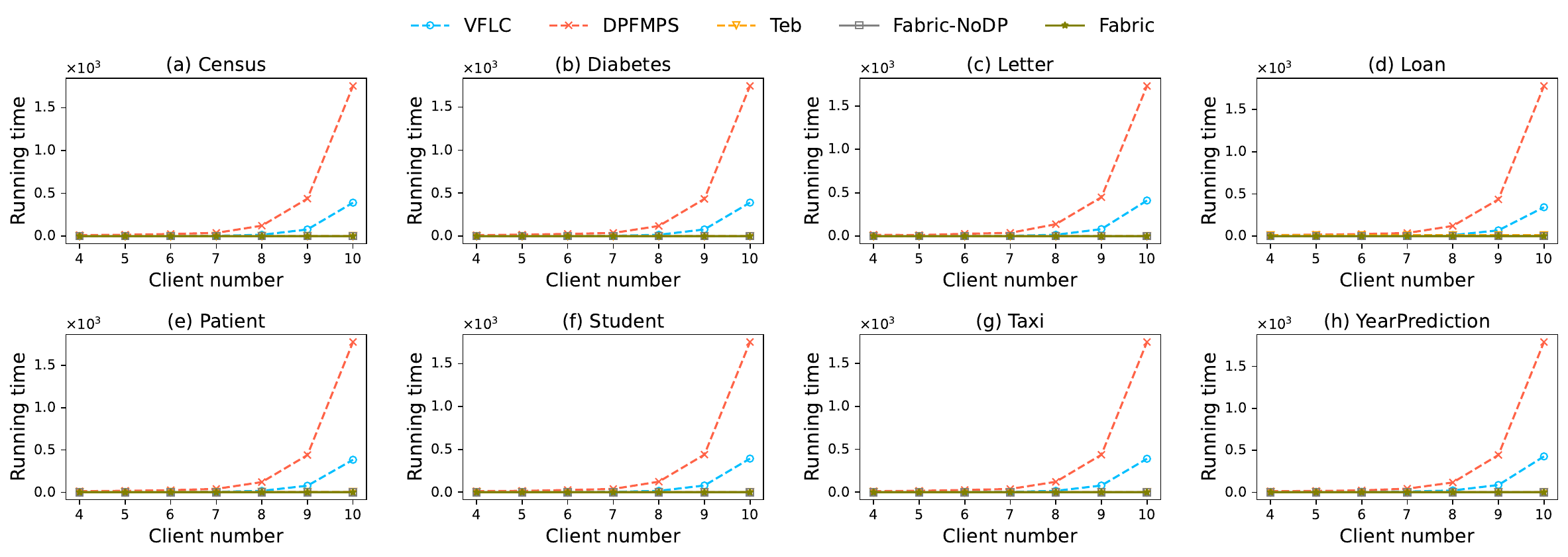}
  \vspace{-2em}
    \captionof{figure}{Runtime with varying numbers of clients across datasets.}
      \vspace{-1em}
    \label{fig:runtime_varying_clients}
\end{figure*}

\section{EXPERIMENTS}\label{sec: experiments}
In this section, we evaluate the effectiveness of the proposed methods by conducting experiments on several real-world datasets and comparing their performance with SOTA methods.
Additionally, we investigate the impact of various parameters on our methods.
To ensure reproducibility, we provide the open-source code \footnote{\textcolor{black}{\href{https://github.com/whu-totemdb/Fabric.git}{https://github.com/whu-totemdb/Fabric.git}}} used in our experiments.

\subsection{Settings}

\myparagraph{Datasets}
We evaluate our methods and SOTA methods on ten widely used real-world datasets~\cite{Bera19Fair,burrows2017incidence,HMDA2017}, including \Census \cite{Bera19Fair}, \Diabetes \cite{burrows2017incidence}, \Letter \cite{slateletter}, \Loan  \cite{2017loan}, \Patient \cite{strack2014impact}, \Student \cite{Kuzilek2015open}, \Taxi \cite{meg2017taxi}, \YearPredictionMSD \cite{2017year_prediction_msd}, \seconddata \cite{Bera19Fair} and \HMDA \cite{HMDA2017}.
Table~\ref{table:Dataset Table} provides detailed statistics of the ten datasets used in our experiments. 
These datasets cover diverse domains and scales, ranging from tens of thousands to millions of instances.
The first eight datasets are used to evaluate clustering effectiveness, parameter sensitivity, robustness to feature heterogeneity, and scalability with respect to the number of clients.
They are drawn from prior work on balanced clustering and federated clustering benchmarks \cite{Ding16KMeans,li2022differentially,Bera19Fair}.
In these datasets, \seconddata and \HMDA contain millions of records and are used to test \Fabric's computational efficiency. 
For each dataset, we sample a subset of instances and features. 

\myparagraph{Partitions}\label{partitioning} 
To simulate a realistic federated environment while maintaining experimental consistency, we consider two partitioning schemes: feature-balanced and feature-skewed.
In the feature-balanced setting, each client is assigned the same number of features.
In the feature-skewed setting, clients are assigned unequal numbers of features, leading to heterogeneous local datasets.
We use the coefficient of variation (CV) to quantify feature imbalance, where a CV of 0 indicates perfect balance and larger values indicate greater skewness.
By adjusting feature allocation to match the target CV values, we generate distributions ranging from fully balanced to highly skewed, enabling a systematic evaluation of method performance across varying levels of feature heterogeneity.

\myparagraph{Parameter settings}
In the federated setting, our experiments involve a central server and multiple clients, each holding a local partition of the data. 
We conduct experiments with four clients by default, but the proposed method is designed to scale seamlessly to any number of clients without performance loss.
For hyperparameter tuning, we evaluate several candidate values: the penalty parameter $\rho$ is selected from $\{0.005, 0.01, 0.02, 0.05, 0.1\}$, the privacy budget $\epsilon$ from $\{0.5, 1, 2, 5, 8, 10\}$, and the block size $n_b$ is tuned over the range $[4, 1024]$.
The balance bounds are set to $\alpha_j=\max\{1,\lfloor0.95n/k\rfloor\}$ and $\beta_j=\lceil1.05n/k\rceil$ for all clusters by default, while our method can be readily extended to other balance bounds.
We set the hyperparameters of the other methods as recommended in their original papers.
We ran all experiments 10 times and reported the average results.

\myparagraph{Metrics} We use several commonly used metrics to evaluate utility and balance. 
The sum of squared error (SSE) and objective function value (Obj), which are given by the value of \eqref{eq: fabric means Lagrangian} and \eqref{eq:fabric means DP Lagrangian}, are employed to assess clustering utility. 
SSE is used for comparison with other methods, while Obj is used to evaluate the impact of parameters.
For balance evaluation, we use the standard deviation of cluster sizes (SDCS) \cite{Lin19BalancedClustering}, which is defined as follows:
\begin{align}\label{eq: SDCS}
\small
SDCS = \sqrt{\frac{1}{k-1} \sum_{i=1}^{k} \left(n_i - \frac{n}{k}\right)^2},
\end{align}
where $k$ is the total number of clusters, $n_i$ is the number of data points in cluster $i$, and $n$ is the total number of data points. 
A higher SDCS value indicates poorer balance.
To assess the efficiency of the methods, we also use runtime as a metric.

\myparagraph{Implementations}
All experiments are conducted on a Windows 11 Professional workstation with an Intel Core i9-14900KF processor, 128 GB RAM, and a 2 TB SSD. 
We implement our algorithms in MATLAB R2022a.

\myparagraph{Baselines}
We compare our proposed methods with seven baselines: \Lloyd, \CDKM~\cite{nie2022CDKM}, \BCLS~\cite{Liu17Balanced}, \FCE~\cite{zhou2024fair}, \VFLC~\cite{Ding16KMeans}, \DPFMPS~\cite{li2022differentially}, and \Tebmean~\cite{ji2025federated}. 
\CDKM and \Lloyd are two $k$-means clustering without balance constraints. 
\BCLS and \FCE represent current SOTA balanced clustering methods.
\VFLC, \DPFMPS, and \Tebmean are VFL-based clustering methods.
For a fair comparison, the same initialization is used, with the same initial $\bF$ when required.

\subsection{Effectiveness Analysis}
We evaluate the effectiveness of \Fabric on the eight effectiveness datasets in terms of the utility-balance trade-off, parameter sensitivity, and robustness to feature heterogeneity. Specifically, we conduct experiments to:
(i) study how $SSE$ and $SDCS$ change with different values of $k$;
(ii) investigate the effects of the privacy budget $\epsilon$ and the penalty parameter $\rho$;
(iii) analyze how $Obj$ varies with CV to assess the performance under feature heterogeneity.

\myparagraph{Impact of varying $k$ on $SDCS$}
Fig.~\ref{fig: Balance1} compares the balance performance under different values of $k$.
Notably, \Lloyd, \CDKM, \VFLC, and \DPFMPS do not incorporate balance constraints, resulting in significantly higher $SDCS$ than methods that explicitly enforce balance.  
In contrast, balance-aware clustering methods yield cluster sizes that are more consistent with the target distribution, among which \FCE and \Fabric achieve the best overall balance performance.
Although enforcing differential privacy slightly compromises balance quality, \Fabric still surpasses \BCLS and \Tebmean on most datasets with respect to balance.

\myparagraph{Impact of varying $k$ on $SSE$}
Fig.~\ref{fig: SSE1 at varying k} compares $SSE$ across methods for varying $k$ on representative datasets. 
It can be observed that as the number of clusters increases, the $SSE$ of all methods decreases. 
Due to the absence of balance constraints, \CDKM and \Lloyd exhibit the lowest $SSE$. 
For most values of $k$, \FabricNoDP achieves lower $SSE$ compared to \BCLS, \FCE, \VFLC and \Tebmean across various datasets.
Furthermore, the $SSE$ attained by \FabricNoDP is often close to that of \CDKM, highlighting not only its inheritance of \CDKM’s local convergence capability but also its seamless and lossless adaptation to the VFL setting.
In addition, \Fabric consistently outperforms  \DPFMPS in terms of $SSE$, as it avoids injecting noise into cluster centers.

\myparagraph{Parameter sensitivity}
Fig.~\ref{fig:pareto_rho} visualizes the trade-off between clustering quality and balance by plotting the final solutions of \Fabric in the ($SSE$, $SDCS$) plane as $\rho$ increases. 
Across datasets, increasing $\rho$ consistently shifts the solution toward lower $SDCS$ (better balance) at the cost of higher $SSE$. 
This is because a smaller $\rho$ relaxes the balance constraint, giving the algorithm more flexibility to find a solution with a lower $SSE$, potentially at the expense of cluster balance.
Based on the sensitivity analysis, $\rho$=0.01 offers a robust trade-off on most datasets and is used by default in subsequent experiments.
We additionally report the convergence curves in Fig.~\ref{fig: Fabric-means SSE at varying rho} and Fig.~\ref{fig: Fabric-meansdp SSE1 at varying rho}.
We observe that as $\rho$ decreases, \Fabric converges faster. 
For a fixed training loss, a smaller  $\rho$ reaches the target in fewer iterations.
\FabricNoDP demonstrates the same behavior.

Fig. \ref{fig: Fabric-means-dp SSE1 at varying epsilon} shows that \Fabric converges stably across different values of $\epsilon$. 
The results also reflect the privacy-utility trade-off: smaller $\epsilon$ introduces more noise and provides stronger privacy, but reduces clustering utility.

\begin{table*}[]
\small
\caption{Comparison in terms of running time across different sample sizes and datasets (seconds). MLE indicates instances where the memory exceeds the device's capacity, while SLE denotes cases where the sample size exceeds the dataset size.}
\vspace{-0.6em}
\centering
\label{tab: running time}
\resizebox{0.88\textwidth}{!}{
\begin{tabular}{ccccccccccc}
\toprule
\textbf{Dataset} & \textbf{Method} & 10K & 20K & 50K & 0.1M & 0.2M & 0.5M & 1M & 2M & 5M \\ 
\midrule
\multirow{6}{*}{\HMDA}      
& \BCLS     & 34.5 & 136.8 & 733.8 & MLE & MLE & MLE & MLE & MLE & MLE \\
& \FCE      & 117.8 & 264.3 & 665.6 & MLE & MLE & MLE & MLE & MLE & MLE \\
& \VFLC     & 105.4 & 206.5 & 585.7 & 1073.3 & 2389.0 & 5807.5 & 13121.5 & MLE & MLE \\
& \DPFMPS   & 161.1 & 718.0 & MLE & MLE & MLE & MLE & MLE & MLE & MLE \\
& \Tebmean   & 3.5 & 7.2 & 20.5 & 43.3 & 87.6 & 746.9 & 1571.3 & 5517.3 & 15314.5\\
& \FabricNoDP   & \textbf{2.5} & \textbf{5.3} & 14.3 & \textbf{31.3} & \textbf{70.9} & \textbf{230.8} & \textbf{608.5} & \textbf{1866.3} & 9541.2 \\
& \Fabric & 3.2 & 6.2 & \textbf{10.3} & 32.9 & 103.7 & 262.0 & 675.8 & 2161.1 & \textbf{9187.5} \\
\midrule
\multirow{6}{*}{\seconddata} 
& \BCLS     & 37.4 & 140.5 & 720.1 & MLE & MLE & MLE & MLE & MLE & SLE \\
& \FCE      & 127.9 & 277.6 & 678.5 & MLE & MLE & MLE & MLE & MLE & SLE \\
& \VFLC     & 82.4 & 162.2 & 410.4 & 899.7 & 1732.9 & 5247.8 & 11091.5 & MLE & SLE \\
& \DPFMPS   & 249.0 & 1244.6 & MLE & MLE & MLE & MLE & MLE & MLE & SLE \\
& \Tebmean   & 3.8 & 7.4 & 21.6 & 49.3 &93.1 & 843.1 & 1627.3 & 3384.9 & SLE \\
& \FabricNoDP   & \textbf{2.1} & \textbf{4.7} & 12.1 & \textbf{27.3} & \textbf{64.5} & \textbf{220.0} & \textbf{602.9} & 1878.5 & SLE \\
& \Fabric & 2.8 & 6.7 & \textbf{11.3} & 41.1 & 72.1 & 267.6 & 652.2 & \textbf{1784.0} & SLE \\
\bottomrule
\end{tabular}
}
\vspace{-1.2em}
\end{table*}

\myparagraph{Robustness to feature heterogeneity}
Fig. \ref{fig: SSE at varying CV} highlights the remarkable robustness of our methods to CV. 
In contrast to SOTA methods that show significant $SSE$ fluctuations, our approach remains highly stable. 
Specifically, \ FabricNoDP's performance is theoretically invariant to feature distribution, as our block matrix decomposition strategy yields a solution equivalent to the centralized formulation. 
Its private counterpart, \Fabric, also achieves nearly identical $SSE$, with slight variations attributable to DP noise and iterative optimization. 

\subsection{Efficiency Analysis}
We evaluate the efficiency of our methods from two aspects: 
(i) running time and 
(ii) communication efficiency.

\begin{table}[t]
\vspace{-0.8em}
\caption{Communication rounds of \Fabric at varying $n_b$.}
\vspace{-0.6em}
\label{tab:rounds_n_b}
\centering
\begin{tabular}{c|cccccc}
\toprule
\textbf{$n_b$} & $2^0$ & $2^2$ & $2^4$ & $2^6$ & $2^8$ & $2^{10}$ \\
\midrule
Rounds &  1.0M   & 0.3M & 62.5k & 15.6k & 3.9k & 1.0k \\
\bottomrule
\end{tabular}
\vspace{-2.2em}
\end{table}

\myparagraph{Running time} 
Table~\ref{tab: running time} reports the runtime for various methods on two large-scale datasets with varying sample sizes. 
\FabricNoDP and \Fabric achieve a 26$\times$ average speedup in runtime over the baselines.
The runtime differences stem from the algorithms' computational complexity. 
Moreover, \BCLS, \FCE, and \DPFMPS suffer from MLE issues as dataset sizes grow, whereas \FabricNoDP and \Fabric scale efficiently, making them suitable for large-scale deployments.

Fig.~\ref{fig:runtime_varying_clients} further demonstrates how the number of clients, varying from 4 to 10, affects the runtime of VFL methods across all datasets included in this experiment.
As the number of clients increases, the runtime of \VFLC and \DPFMPS rises sharply, since both methods rely on constructing weighted grids with computational complexity $\mathcal{O}(k^H)$.
In contrast, the runtime of \FabricNoDP, \Fabric, and \Tebmean remains relatively stable as the number of clients increases, demonstrating their strong scalability.
For \FabricNoDP and \Fabric, this scalability is attributable to the negligible additional computational overhead introduced by the decomposition strategy.

\myparagraph{Communication efficiency}
Table~\ref{tab:rounds_n_b} presents the communication rounds required by \Fabric under different block sizes $n_b$, based on 10 trials with $k = 4$, $\rho = 0.01$, and 100 iterations on 5,000 randomly sampled instances.
The results reveal a dramatic reduction in communication overhead as $n_b$ increases.
Specifically, when $n_b = 2^{10}$, \Fabric requires only 1.0k communication rounds, representing a 1000 $\times$ reduction compared to when $n_b=1$.
This substantial improvement stems from our block update strategy, which processes multiple data points per communication round.

\section{Conclusions}
This paper presented \Fabric, a federated balanced clustering algorithm with formal privacy guarantees for VFL.
\Fabric addresses the key challenge of jointly achieving clustering balance and privacy under heterogeneous feature distributions.
Extensive experiments across ten real-world datasets demonstrated that \Fabric achieves superior utility-balance trade-offs, with up to 1000$\times$ reductions in communication rounds and an average 26 $\times$ speedup in runtime.
Future work will extend our framework to settings with dynamic client participation.
Moreover, we will investigate pruning-based methods to further accelerate clustering.


\section{Acknowledgment}
This work was supported by the National Key R$\&$D Program of China (2023YFB4503600), the National Natural Science Foundation of China (62202338, 62372337, 62372340, and 62662058), the Key R$\&$D Program of Hubei Province (2023BAB081), the Major Technical Research Project of Hubei Province (2023BAA018), and the Hubei Key Laboratory of Big Data in Science and Technology (No.~E6KF001001).

\section{AI-Generated Content Acknowledgement}
We used ChatGPT only to check grammar in the manuscript. The authors completed all technical ideas, algorithmic designs, experimental analyses, manuscript writing, and code.




\bibliographystyle{IEEEtran}
\bibliography{sample-base}

\vfill
\end{document}